\documentclass[letterpaper, 10 pt, conference]{ieeeconf}
\IEEEoverridecommandlockouts
\usepackage{cite}
\usepackage{amsmath,amssymb,amsfonts}
\usepackage{algorithmic}
\usepackage{graphicx}
\usepackage{gensymb}
\usepackage[hidelinks]{hyperref}
\usepackage{textcomp}
\usepackage{xcolor}
\usepackage{dblfloatfix}
\def\BibTeX{{\rm B\kern-.05em{\sc i\kern-.025em b}\kern-.08em
    T\kern-.1667em\lower.7ex\hbox{E}\kern-.125emX}}
\begin{document}

\title{\LARGE \bf eViper: A Scalable Platform for Untethered Modular Soft Robots\\
}
\author{
Hsin Cheng, Zhiwu Zheng, Prakhar Kumar, Wali Afridi, Ben Kim\\ Sigurd Wagner, Naveen Verma, James C. Sturm and Minjie Chen\\
{\it Princeton University, Princeton NJ, 08540}
}
\bstctlcite{IEEEexample:BSTcontrol}
\maketitle
\thispagestyle{empty}
\pagestyle{empty}

\begin{abstract}
    Soft robots present unique capabilities, but have been limited by the lack of scalable technologies for construction and the complexity of algorithms for efficient control and motion. These depend on soft-body dynamics, high-dimensional actuation patterns, and external/onboard forces. This paper presents scalable methods and platforms to study the impact of weight distribution and actuation patterns on fully untethered modular soft robots. An {\it extendable Vibrating Intelligent Piezo-Electric Robot} (eViper), together with an open-source {\it Simulation Framework for Electroactive Robotic Sheet} (SFERS) implemented in PyBullet, was developed as a platform to analyze the complex weight-locomotion interaction. By integrating power electronics, sensors, actuators, and batteries onboard, the eViper platform enables rapid design iteration and evaluation of different weight distribution and control strategies for the actuator arrays. The design supports both physics-based modeling and data-driven modeling via onboard automatic data-acquisition capabilities. We show that SFERS can provide useful guidelines for optimizing the weight distribution and actuation patterns of the eViper, thereby achieving maximum speed or minimum cost of transport (COT).
\end{abstract}


\section{Introduction}\label{sec1}

Research on robots with rigid components has a long history. Designing and controlling soft robots for a variety of tasks – from crawling on surfaces to carrying different payloads – is just beginning to be understood~\cite{Rich2018, Li2022}. Soft robots are attractive because: (1) they are deformable and can closely mimic animal behavior; (2) they are soft and thus more collision-resilient than traditional rigid-body robots; (3) they can perform tasks that rigid robots cannot perform~\cite{Rus2015, Picardi17, iida2011soft, laschi2016soft, wu2019insect, Jafferis2019, elastomer19, footpad21}. The actuators in a soft robot can be activated electrically, thermally, or pneumatically~\cite{el2020soft}. Many factors affect the locomotion of soft robots, including the actuation patterns of the actuators, the weight distribution across the soft robot body, and the capability of untethered operation. Some of the above demonstrations are tethered hence only consider the impact of weight distribution in the context of tethered operation. Some other designs have limited control degrees of freedom (DOF).  Research on the energy efficiency, and actuation-locomotion co-design is still at the early stage~\cite{Efficiency17}.

\begin{figure}[t]
\centering
\includegraphics[width=\columnwidth]{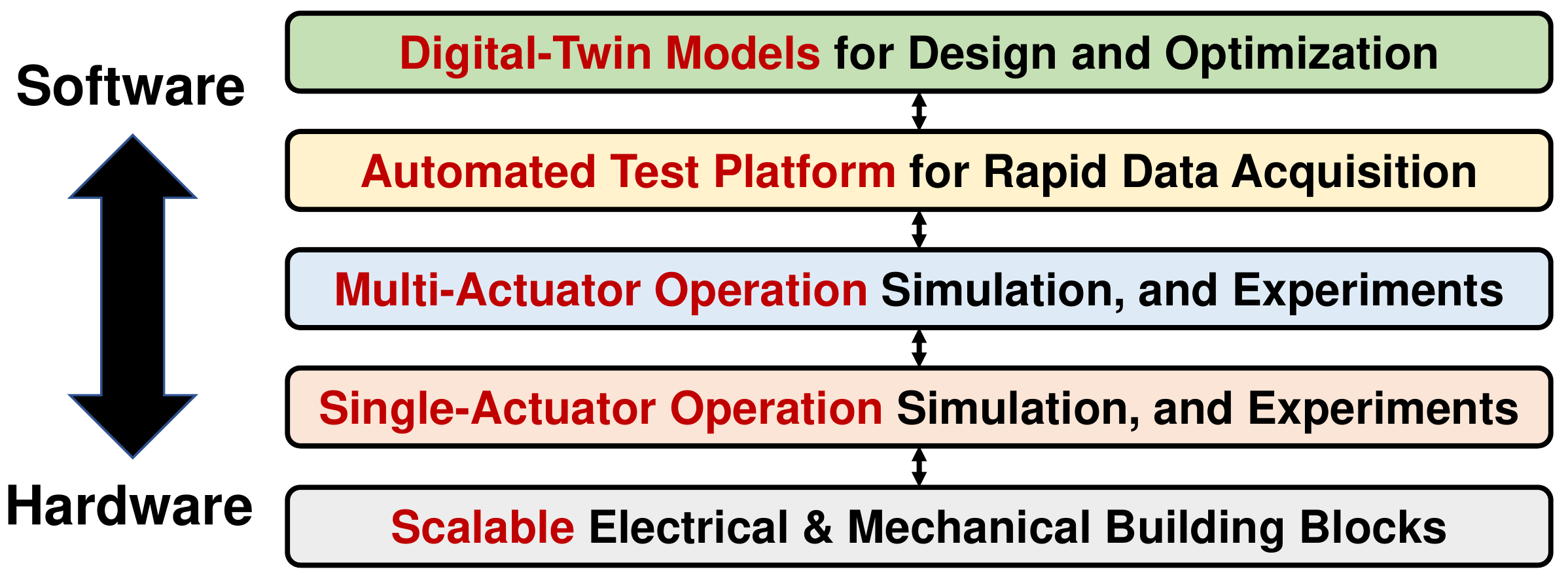}
\vspace{-18pt}
\caption{A systematic approach to studying untethered, modular, scalable soft-robots ranging from basic building blocks, single-actuator, multi-actuator, automatic testing, and data acquisition, to full robot digital twin.}
\vspace{-10pt}
\label{fig:flow}
\end{figure}

\renewcommand{\thefigure}{4}
\begin{figure*}[b]
\centering
\vspace{-10pt}
\includegraphics[width=\textwidth]{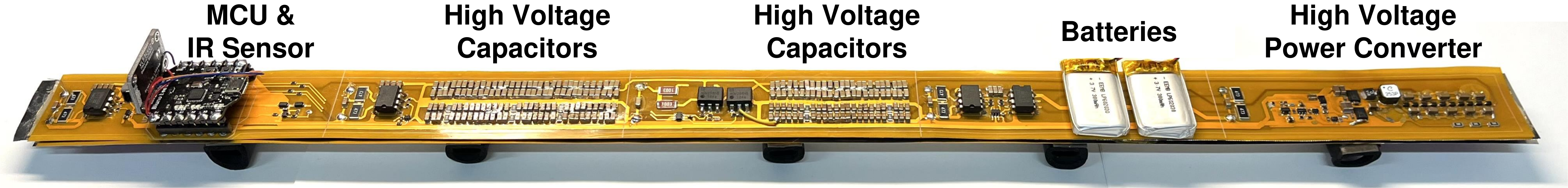}
\vspace{-20pt}
\caption{Photo of a five-actuator eViper.}
\label{fig:FiveActPrototype}
\vspace{-10pt}
\end{figure*}

\renewcommand{\thefigure}{2}
\begin{figure}[!t]
\centering
\includegraphics[width=\columnwidth]{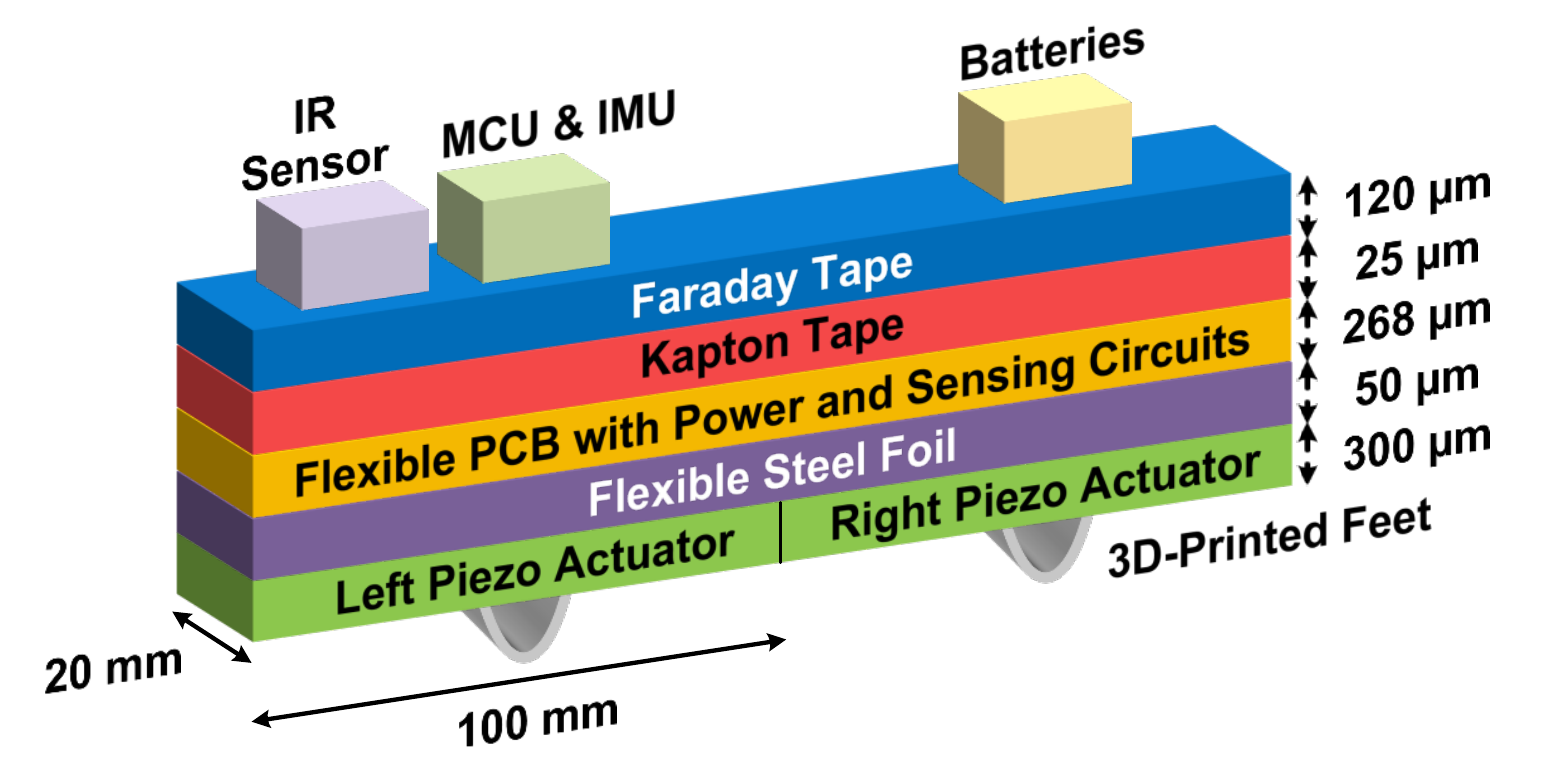}
\vspace{-20pt}
\caption{Mechanical structure of an illustrative two-actuator eViper comprising actuators, power electronics, sensors, microcontroller, and batteries. The actuators and the steel foil are bonded with epoxy glue. Other layers are bonded with double-sided tape or Kapton tape.}
\label{fig:mechanical}
\vspace{-10pt}
\end{figure}

\renewcommand{\thefigure}{3}
\begin{figure}[t]
\includegraphics[width=\columnwidth]{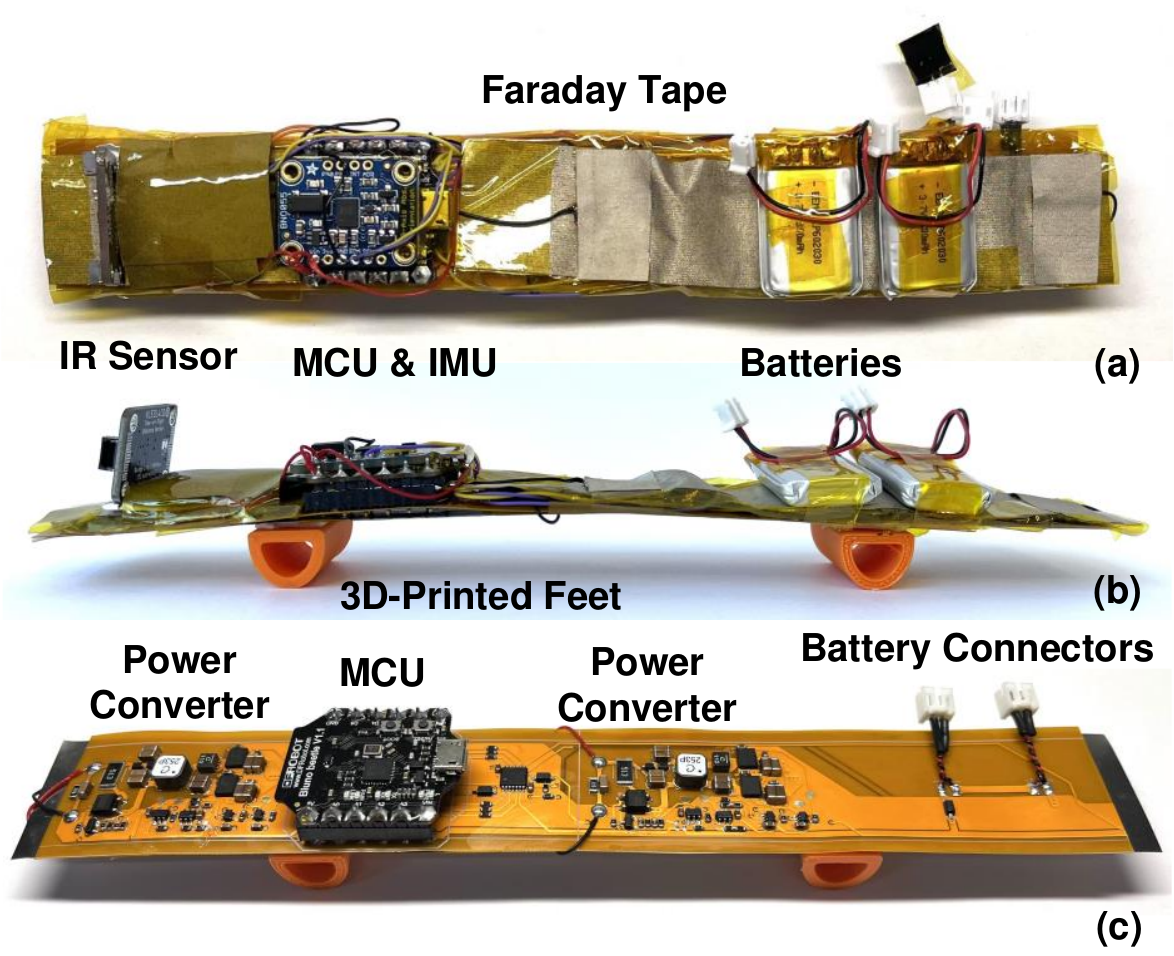}
\vspace{-20pt}
\caption{(a) Top view, (b) Side view, and (c) Embedded power electronics of a two-actuator eViper. The high voltage circuits are shielded by a Kapton insulating tape to ensure electrical safety, and a conductive Faraday tape connected to the negative terminal of the battery to block electromagnetic interference (EMI). The length and width of the robot are 200~mm $\times$ 20~mm. The height of the two 3D-printed robot feet is 10 mm.}
\label{fig:robotTopSideView}
\vspace{-10pt}
\end{figure}


It's challenging to use analytical methods to describe the complex interactions between the soft actuator, the robot body, the weight of electronics and batteries, and the environment. Researchers have explored several aspects of soft robot locomotion, including crawling and jumping~\cite{Zheng2021, ZhiwuRobo, ZhiwuICRA}, turning~\cite{yu2020crawling, wu2018structure}, manipulating friction~\cite{ICRA16, calabrese2019soft}, and frequency-dependent behavior~\cite{MEMS19, Hassan16}. However, only a few of the reported analytical methods can be applied to highly scalable and complex robots.

Figure~\ref{fig:flow} illustrates the key principles of a systematic approach to studying the impact of weight distribution and actuation patterns on untethered modular multi-actuator soft robot locomotion. The method includes five central aspects from hardware to software: (1) designing modular and scalable mechanical and electrical building blocks; (2) modeling and operating the soft robot actuated with a single actuator; (3) modeling and operating the soft robot actuated with multiple actuators; (4) creating automatic test platform and data acquisition; and (5) establishing digital twin models of the soft robot for design, optimization, and control. This approach can be used to design scalable multi-actuator soft robots for different purposes, such as improving the energy efficiency of locomotion in different environments, or maximizing the payload of the robot for weight carrying.

To demonstrate the effectiveness of the proposed method, we present a fully-untethered {\it extendable Vibrating Intelligent Piezo-Electric Robot} -- the eViper -- together with an open-source {\it Simulation Framework for Electroactive Robotic Sheet} -- the SFERS -- implemented in PyBullet \cite{coumans2016pybullet}. The paper includes the following contributions:
\begin{itemize}
    \item Develops an integrated hardware-to-software platform to study the locomotion of soft robots.
    \item Introduces a hierarchical technique to design untethered, scalable, and modular piezoelectric soft robots.
    \item Customized a lightweight, high voltage power architecture for driving piezoelectric actuators.
    \item Presents detailed simulation and experimental results for the impact of weight distribution and actuation patterns on the locomotion of soft robots. 
\end{itemize}

\section{eViper Hardware Implementation}\label{sec2}
\subsection{Mechanical Platform}\label{MP}

Figure~\ref{fig:mechanical} illustrates the mechanical structure of a two-actuator eViper example. The fabrication process is hierarchical, based on separating the components required for different functionality (actuators, mechanical substrate, power electronics, and control circuitry) across different soft and flexible layers. eViper is a peizoelectric soft robot. Each piezoelectric actuator is a 20-mm-wide, 100-mm-long, and 300-$\mu$m-thick commercial Macro Fiber Composite (MFC) unit from Smart Material Corp.~\cite{Smartmaterial}, bonded to a 50-$\mu$m-thick steel foil. On top of the steel foil, we customized a flexible printed circuit board (PCB) that hosts power converters and control circuitry to drive the actuators. To ensure safe testing and high voltage insulation, the power electronics are enclosed in a shielding layer. The shielding layer comprises a 25-$\mu$m-thick Kapton tape covered by a 120-$\mu$m-thick Faraday tape. The Faraday tape is conductive and is connected to the negative terminal of the battery (the electrical ground of the robot), functioning as a Faraday cage and electromagnetic interference (EMI) shield to block the high voltage. The batteries, microcontroller, communication circuits, and auxiliary sensors (all low-voltage) are left outside. Two 3D-printed plastic ``D"-shaped feet are attached to the bottom side of the robot to lift the robot body and ensure robust ground friction. All these functional layers contribute to the weight distribution and the stiffness of the soft robot. These factors, together with the actuation pattern, jointly affect the locomotion efficiency and the payload capacity. 

Figure~\ref{fig:robotTopSideView}(a) and~\ref{fig:robotTopSideView}(b) show the top view and the side view of the assembled robot. The infrared (IR) distance sensor, microcontroller (MCU), inertial measurement unit (IMU), and batteries are visible from the outside. The high-voltage power electronics are shielded by the Kapton and Faraday tapes. Figure~\ref{fig:robotTopSideView}(c) shows the power electronics inside the shielding. The weight of the robot is 44.5 g. Its remaining payload of about 10 g can accommodate more electronics or batteries. The structure of the proposed platform is highly scalable. It can be extended to a larger actuator array or be arranged to other shapes based on need and application. The multi-layer structure and lamination method for eViper are similar to those used in the fabrication of flexible displays.  Eventually, the manufacture of eVipers may adopt many of the low-cost-per-area fabrication techniques of the display industry. Figure~\ref{fig:FiveActPrototype} shows a longer eViper prototype that is built with five actuators. It is constructed with the same hierarchical philosophy and a similar structure of power electronics. As reported in \cite{https://doi.org/10.48550/arxiv.2207.00658}, the five-actuator prototype can crawl and rotate when the actuators are driven with various frequencies. This paper will focus on presenting the locomotion behaviors of the two-actuator eViper.

\subsection{Electrical Platform}\label{EP}

Figure~\ref{fig:electrical} provides an overview of the electrical architecture of the soft robot. The robot carries two 3.7~V 300~mAh lithium polymer batteries (30 $\times$ 20 $\times$ 6~mm, 6.2~g each), capable of continuously driving the two actuators for about an hour. The two batteries are connected in series to provide a 7.4~V input voltage to the power electronics. A key enabling technology for the fully untethered eViper is a compact, efficient, lightweight high-voltage driver which steps the 7.4~V battery voltage to 300~V and 1500~V to drive the actuators. The DC-DC efficiency is about 70\%. The circuit can be reconfigured to produce 600~V, 900~V, and 1200~V if needed. The high-voltage driver consists of three parts: a series resonant inverter that converts DC input voltage to a high-frequency AC voltage, a transformer that steps up the AC voltage to 300~V, and a half-wave Cockcroft–Walton voltage multiplier that rectifies the AC voltage back to DC voltage and multiplies it. The resonant inverter uses transformer parasitics (the magnetizing inductance) as part of the resonant tank. The voltage conversion ratios of the transformer stage and the Cockcroft–Walton voltage multiplier are 1:40 and 1:5, respectively. Each high voltage driver, shown in Figure~\ref{fig:electrical}(b), is capable of driving an actuator up to 30~Hz with controllable phases and duty ratios. More power electronics details are provided in~\cite{Hsin22}. Each driver can provide more than 2~W power with a weight $<$2~g. 

As reported in \cite{https://doi.org/10.48550/arxiv.2207.00658}, the power converter can efficiently drive M-8514-P1 and M-8514-P2 MFC piezoelectric actuators, which require 1500~V and 300~V for strong actuation. The demonstrated two-actuator eViper in this work uses M-8514-P2 for both actuators. Young’s modulus for the actuator is 30 GPa, and is 190 GPa for the steel foil. When a positive voltage is applied across the MFC terminals, the MFC tends to contract while the steel foil tends to retain its length due to its higher Young’s modulus. As a result, the actuator-steel structure bends concave down, as shown in Fig.~\ref{fig:actBending}.

The eViper is controlled by a microcontroller which can support many sensors. A Bluetooth microcontroller (DFRobot DFR0339) takes charge of actuator control, sensor control, and communication with the PC; an infrared (IR) distance sensor (Adafruit VL53L4CD) measures robot velocity and detects the robot location; a nine-axis inertial measurement unit (IMU, Adafruit BNO055) is used for measuring locomotion; and a DC current sensor (Atnsinc INA219) gauges the power consumption of the untethered robot. The microcontroller and the sensors together consume about 0.8~W. The power stage consumes about 0.4~W. The power consumption of the entire robot is about 1.2~W.

\setcounter{figure}{4}
\renewcommand{\thefigure}{\arabic{figure}}
\begin{figure}[t]
\centering
\includegraphics[width=\columnwidth]{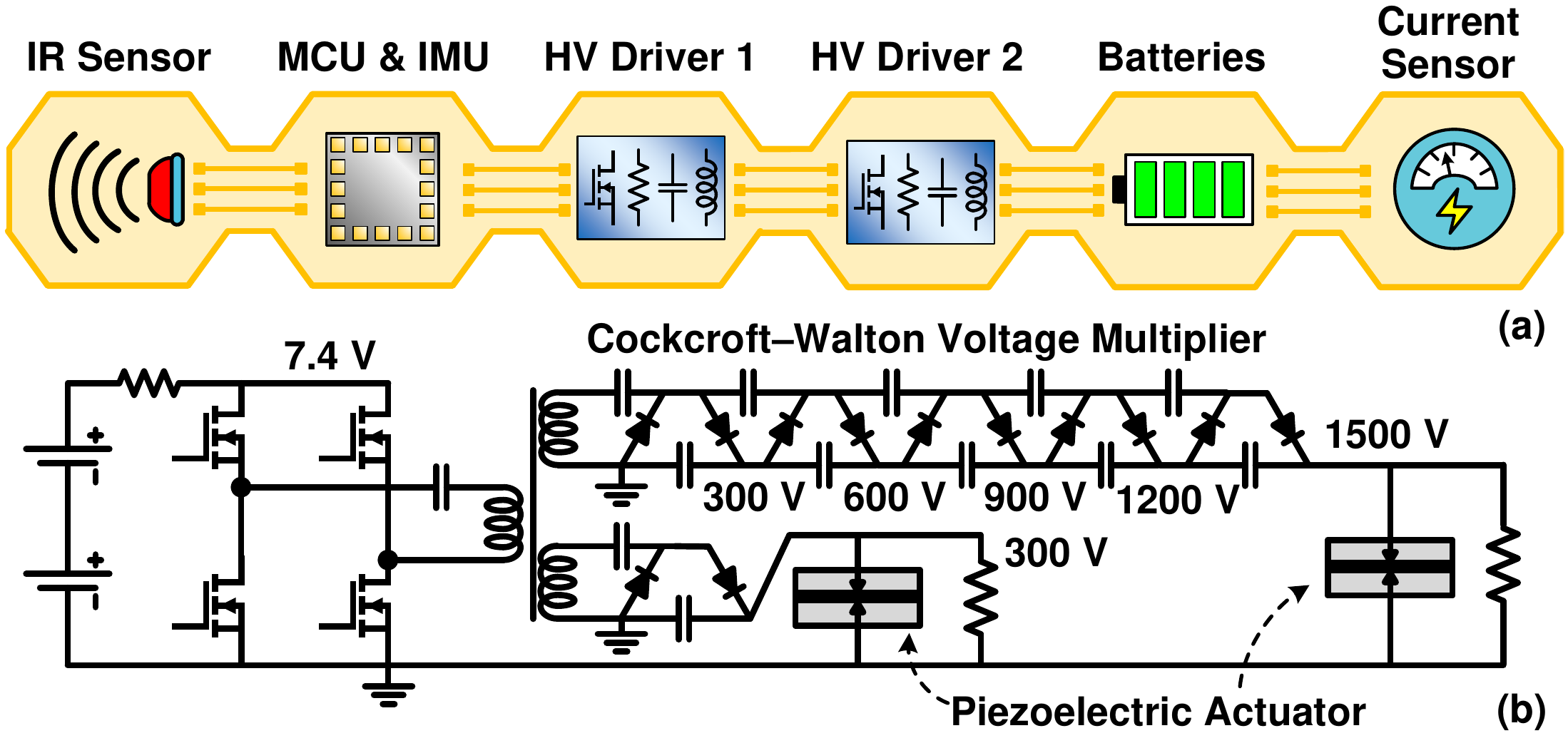}
\vspace{-20pt}
\caption{(a) Block diagram of eViper's electrical platform. (b) The circuit schematic of a high voltage driver with five configurable output voltage options. There are two high voltage drivers on board, and only the 300~V output is used in this work.}
\label{fig:electrical}
\vspace{5pt}
\includegraphics[width=\columnwidth]{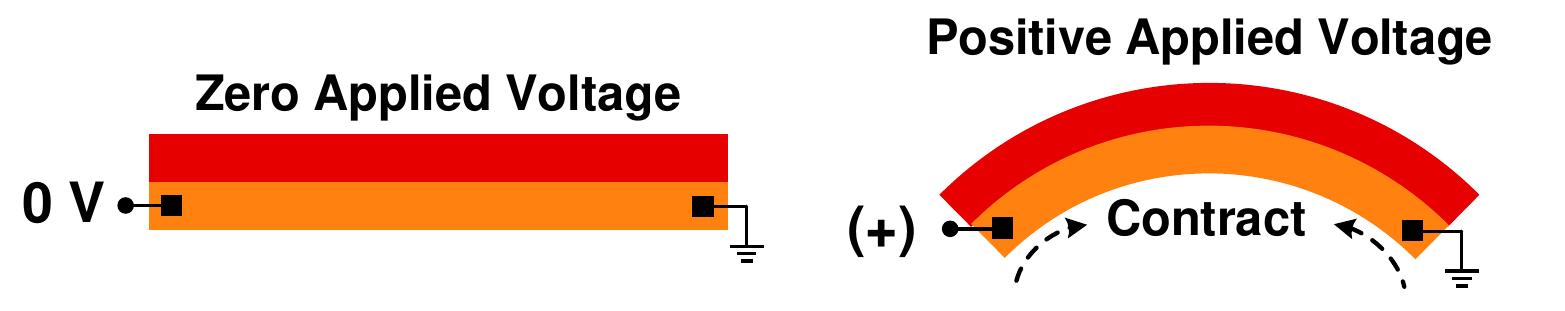}
\vspace{-20pt}
\caption{Bending behavior of the steel-bonded piezoelectric actuator. The red layer is the steel foil, and the orange layer is the piezoelectric actuator.}
\vspace{-10pt}
\label{fig:actBending}
\end{figure}


\section{eViper Operation Mechanism}

\begin{figure}[t]
\centering
\includegraphics[width=\columnwidth]{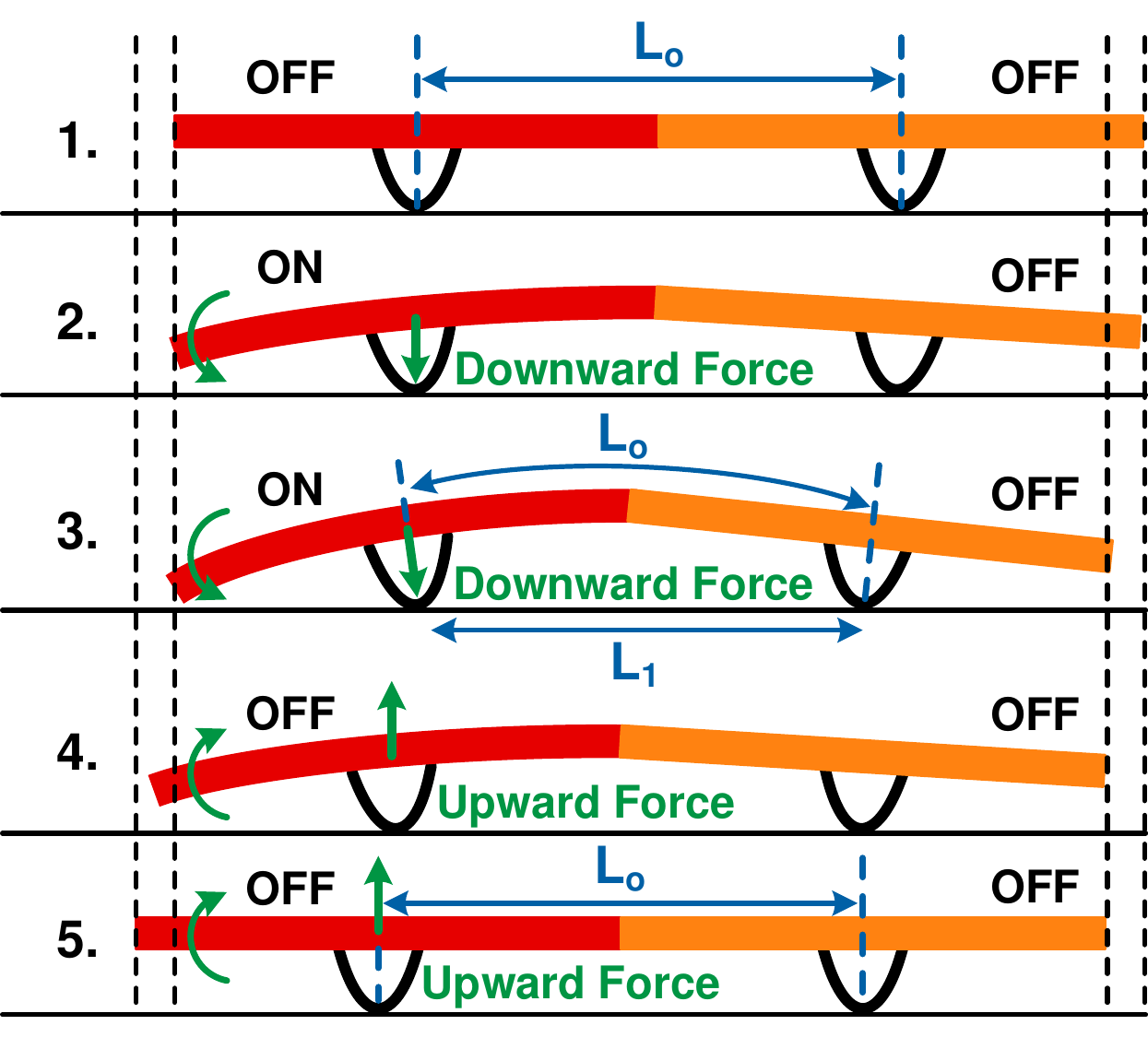}
\vspace{-20pt}
\caption{An illustration depicting the leftward motion when only driving the left actuator. 1. The initial state: $L_o$ is the distance between the two contact points of the feet on the ground. 2. High voltage (300~V) is applied to the left actuator. It starts to bend concave down and applies a downward force to the left foot. The force increases the normal force and the maximum static friction between the left foot and the ground during the dynamic bending process. 3. The left actuator bends to its maximum extent. The distance between the two contact points, $L_1$, must be smaller than the original distance $L_o$ due to the robot curvature. Since the left foot has a higher maximum static friction, the right foot is easier to move and is dragged to the left. 4. 0~V is applied to the left actuator, causing it to flatten and apply an upward force to the left foot, decreasing the normal force and the maximum static friction between the left foot and the ground. 5. In the final step of the cycle, the robot finishes flattening. Since the left foot has a lower maximum static friction than the right foot until the end of the cycle (robot completely flat), the left foot is easier to move and slides to the left. Then the entire robot has moved to the left with respect to its initial state.}
\label{fig:BasicMotionPattern}
\vspace{-10pt}
\end{figure}

\begin{figure}[t]
\centering
\includegraphics[width=\columnwidth]{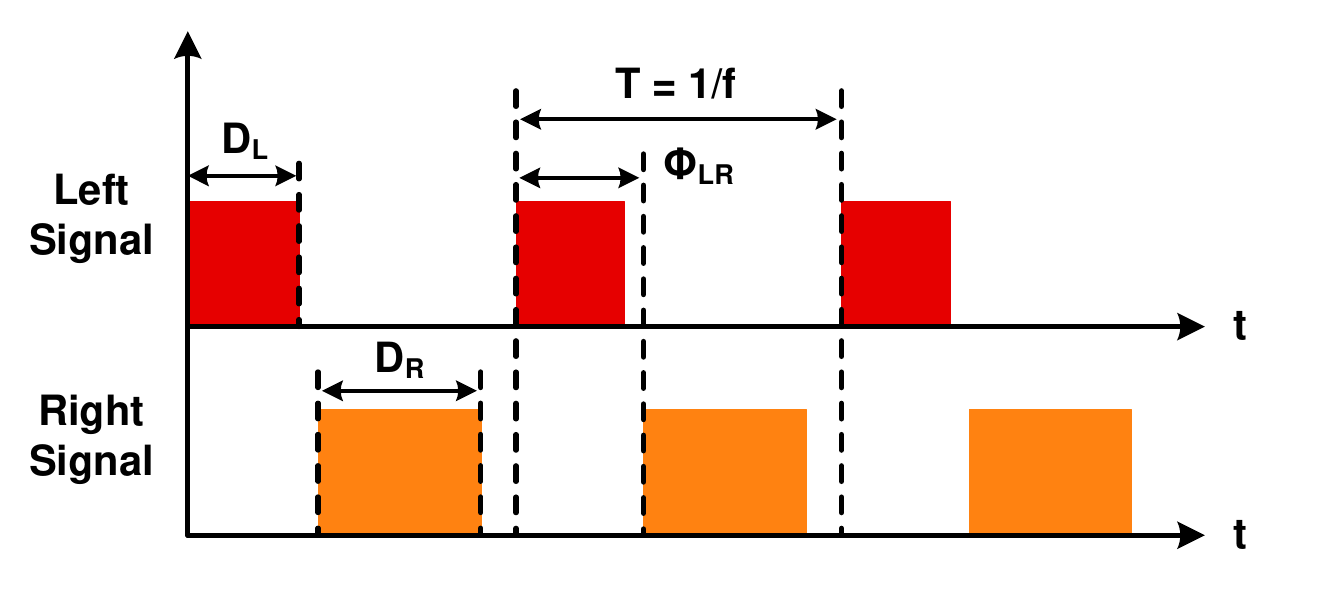}
\vspace{-20pt}
\caption{Example of periodic control signals when driving both actuators with a specific frequency ($f$), phase ($\Phi_{LR}$), duty ratio of the left actuator ($D_{L}$), and duty ratio of the right actuator ($D_{R}$). Frequency is the reciprocal of the signal period $T$. Phase is the fraction of a period $T$ between the rising edges of the two signals and is often expressed in degrees (0$\degree$ to 360$\degree$). The duty ratio, when expressed as a percentage, is the fraction of a period $T$ that a signal remains at the high level.}
\label{fig:IntroOfPhaseDuty}
\vspace{-10pt}
\end{figure}


We start analyzing the locomotion of the two-actuator eViper by only driving a single actuator at a time. Motion in the long direction of the robot is created due to actuation asymmetry, weight, and structure. Figure~\ref{fig:BasicMotionPattern} illustrates five general locomotion steps when only driving the left actuator, and we assume the weight distribution across the robot body is uniform. The same principle can also be applied if we only drive the right actuator of an uniform-weight eViper.

While the precise locomotion arises from more complex deformations and forces, the understanding explained in Figure~\ref{fig:BasicMotionPattern} provides an effective starting point for design and trade-off exploration. The effectiveness of locomotion depends on the frequency, strength, and duration of the actuation. Asymmetric weight distribution and asymmetric actuation lead to asymmetric locomotion.

Understanding single-actuator actuation serves as the basis of understanding multi-actuator actuation. The locomotion becomes more complicated when two actuators are operating at various frequencies, duty ratios, and phases, and if weight distribution is asymmetric. Figure~\ref{fig:IntroOfPhaseDuty} shows an example of periodic control signals for both actuators. The enormous control DOF makes it complicated and impractical to develop an analytical model for describing the locomotion. Instead, we build a simulation toolkit to assist in the design and control of the modular soft robot.


\section{SFERS: Simulation Framework for Electroactive Robotic Sheet}\label{SFERS}

\begin{figure}[t]
\centering
\includegraphics[width=\columnwidth]{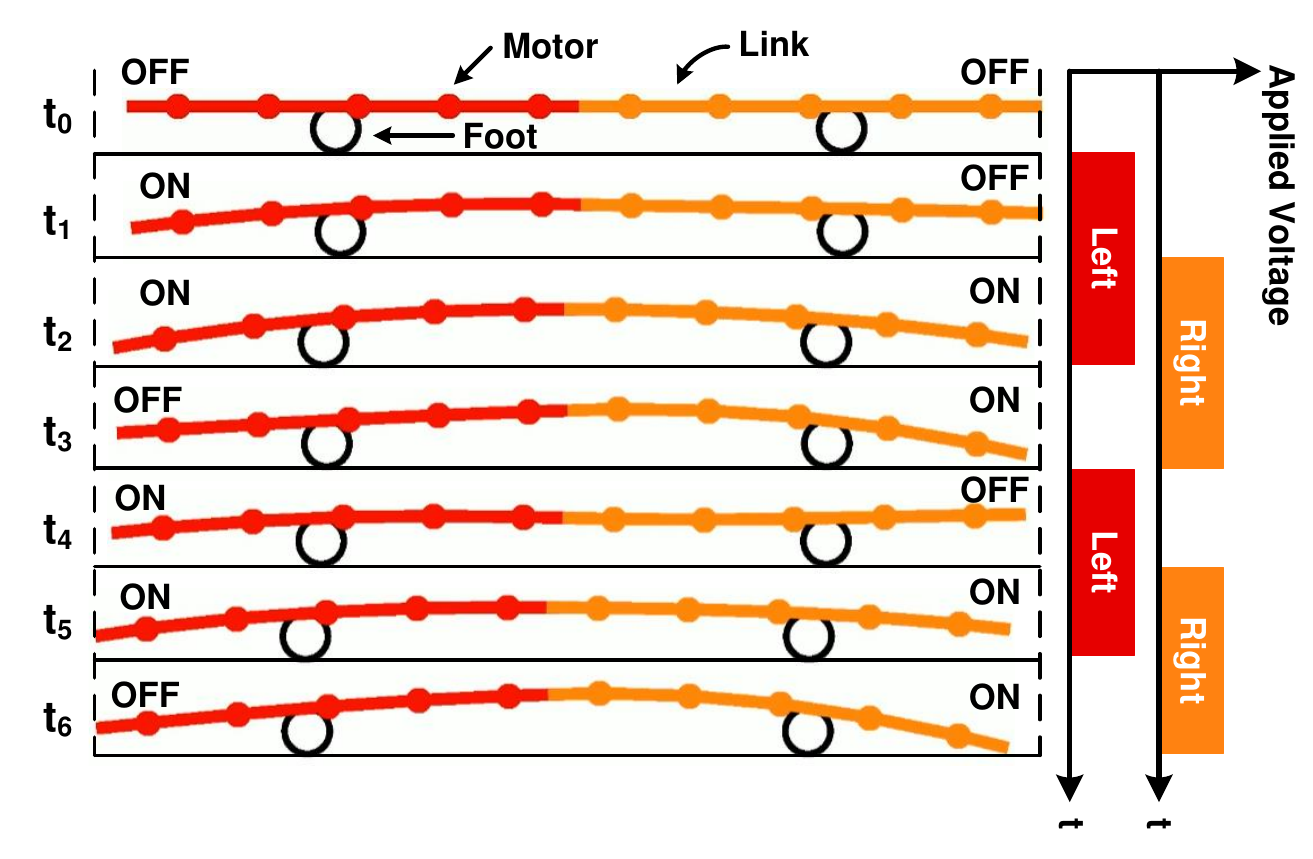}
\vspace{-20pt}
\caption{Illustration showing the concept of the motor-link structure together with several frames ($t_0$ to $t_6$) in a simulation video. We use five motors to simulate one actuator. The time difference between each frame is 20 ms. The right side shows the applied voltage versus time. The red and orange voltages correspond to the left and right actuator, respectively.}
\label{fig:link-torque-model}
\vspace{10pt}
\includegraphics[width=\columnwidth]{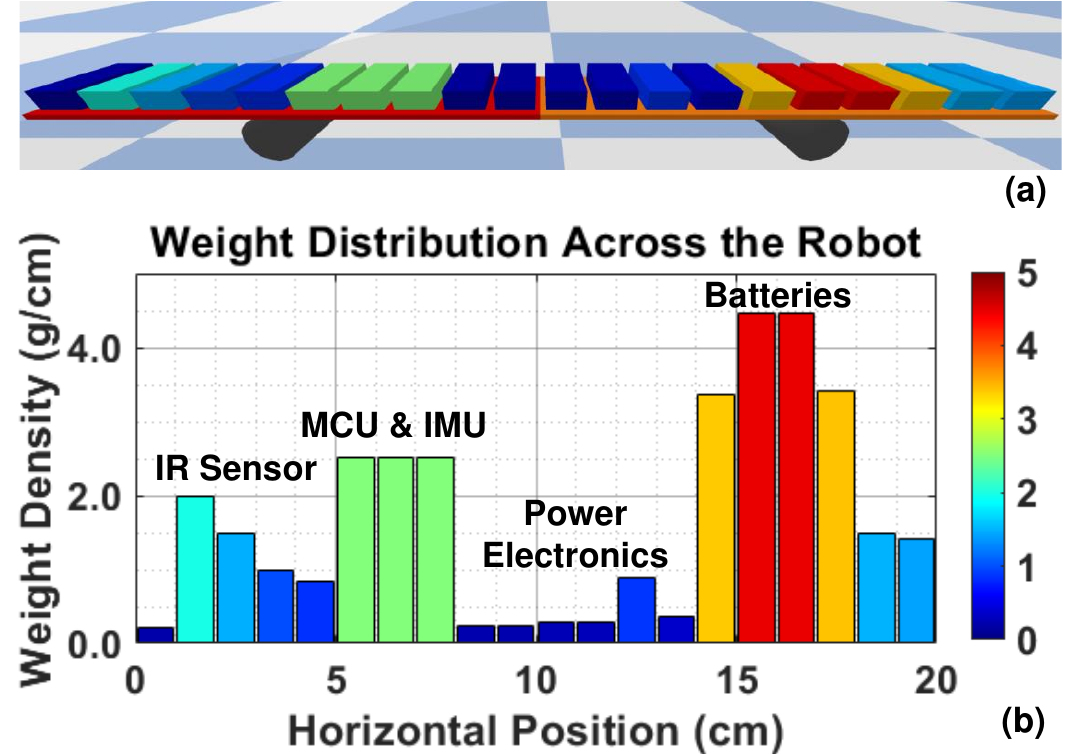}
\caption{(a) The side view of the robot in PyBullet physics engine. There are three layers, from top to bottom: the boxes which represent the weight distribution, the actuators, and the feet of the robot. (b) The estimated weight distribution of the components on top of the robot body. Each bar in this graph is assigned the same color as the corresponding box in Figure~\ref{fig:WeightDistribution}(a).}
\label{fig:WeightDistribution}
\vspace{-10pt}
\end{figure}

We introduce SFERS~\cite{ZhiwuICRA} as an open-source software platform (at {https://github.com/zhiwuz/sfers}) based on PyBullet, for rapid simulation of a multi-actuator soft robot. As shown in Fig.~\ref{fig:link-torque-model}, SFERS models piezoelectric actuators as a series of short and rigid links connected with torque-controlled motors. It computes motor torques from the voltages applied to the actuators and uses the information to drive the soft robot. The model has proved useful for modeling static and dynamic motion of multi-actuator piezoelectric 2-D soft robots~\cite{ZhiwuICRA}. In this work, we model each actuator as six links and five motors. The entire two-actuator robot has eleven links and ten motors (the links at the junction of two actuators are treated as one link). Based on the measured material properties of eViper, the motors in the simulation are set to have a torsional Hooke's constant of 0.32~N$\cdot$m.
In SFERS we model the robot feet as a pair of hard cylinders which are attached to the robot body.

\begin{figure}[t]
\centering
\includegraphics[width=\columnwidth]{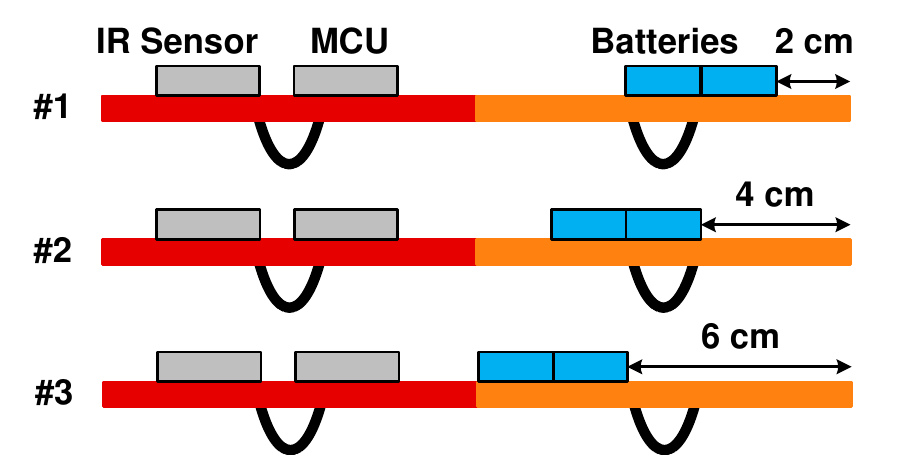}
\caption{The three different battery positions we used to study the impact of weight distribution. We define the figures from top to bottom as battery position \#1, \#2, and \#3, respectively. Battery position \#1 is the default position used in the majority of tests presented in this paper.}
\label{fig:BatPosAll}
\vspace{-10pt}
\end{figure}

The numerous electronic components and batteries are modeled as inflexible boxes attached to the links, as illustrated in Fig.~\ref{fig:WeightDistribution}(a). The weight of each box is different, and the weight assignment is based on the weight distributed over each centimeter along the robot, which is shown in Fig.~\ref{fig:WeightDistribution}(b). The interface between the weight boxes and the robot body is firm and non-flexible. The robot weighs 44.5~g, including the two batteries weighing 12.4~g ($\sim$28\% of the total weight).


The SFERS platform helps to develop a qualitative understanding of robot locomotion. It is a tool to assist more sophisticated soft robot design before experimental prototyping. However, SFERS has many limitations which may prevent it from making a precise prediction of the soft robot locomotion. These limitations include:
\begin{itemize}
    \item SFERS assumes that all functional layers are tightly bonded to each other, which may not be true.
    \item Addition of the electronics and batteries may change the flexibility and the effective Young's modulus.
    \item The interface between the robot feet and the external environment is different from the models.
    \item Non-idealities in the power electronics, e.g., the voltages generated by the power electronics are not ideal pulse-width-modulated square waves but have finite rising and falling time.
    \item The mechanical characteristics of the soft robot may not be uniform across the entire robot. 
    \item The asymmetry across the width is not captured.
\end{itemize}

All these factors may influence the accuracy of the PyBullet model and may introduce discrepancies in robot behavior between simulation and experiment. Sensors and closed-loop control are needed for precise robot manipulation and locomotion. Nevertheless, we found that the scalable SFERS modeling framework can capture the general robot behavior across a wide range of operating frequency, phase, and duty ratios. It provides useful insights on the soft robot locomotion when driving the robot with different actuation patterns and different weight distribution across the robot body.

\section{Experimental Results}\label{ExpResults}

\begin{figure}[t]
\centering
\includegraphics[width=\columnwidth]{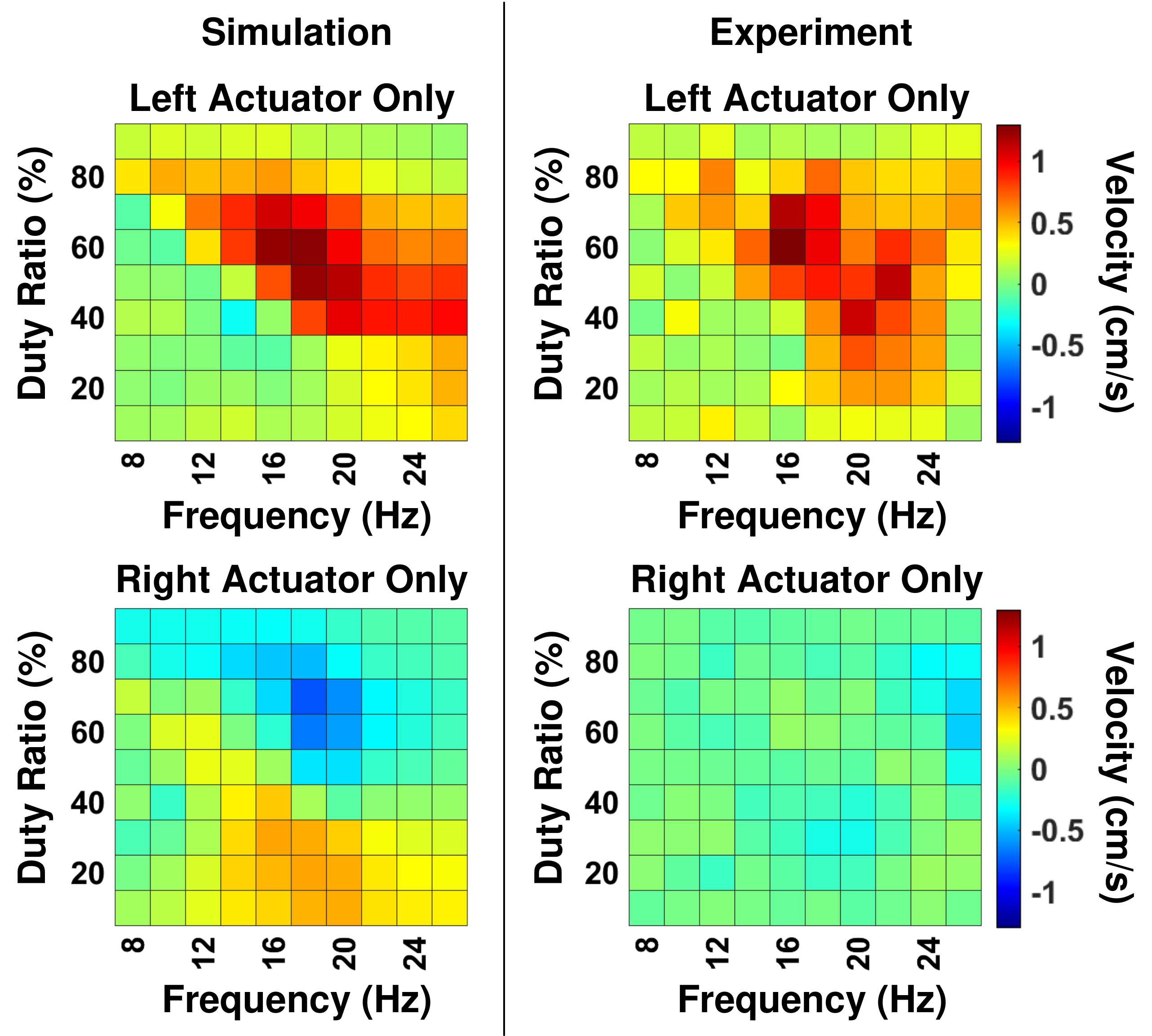}
\vspace{-18pt}
\caption{Simulated (left column) and experimentally measured (right column) robot velocity as a function of frequency and duty ratio when driving only the left actuator (top row) or the right actuator (bottom row). Batteries at position \#1.}
\label{fig:BatPos1}
\vspace{10pt}
\includegraphics[width=\columnwidth]{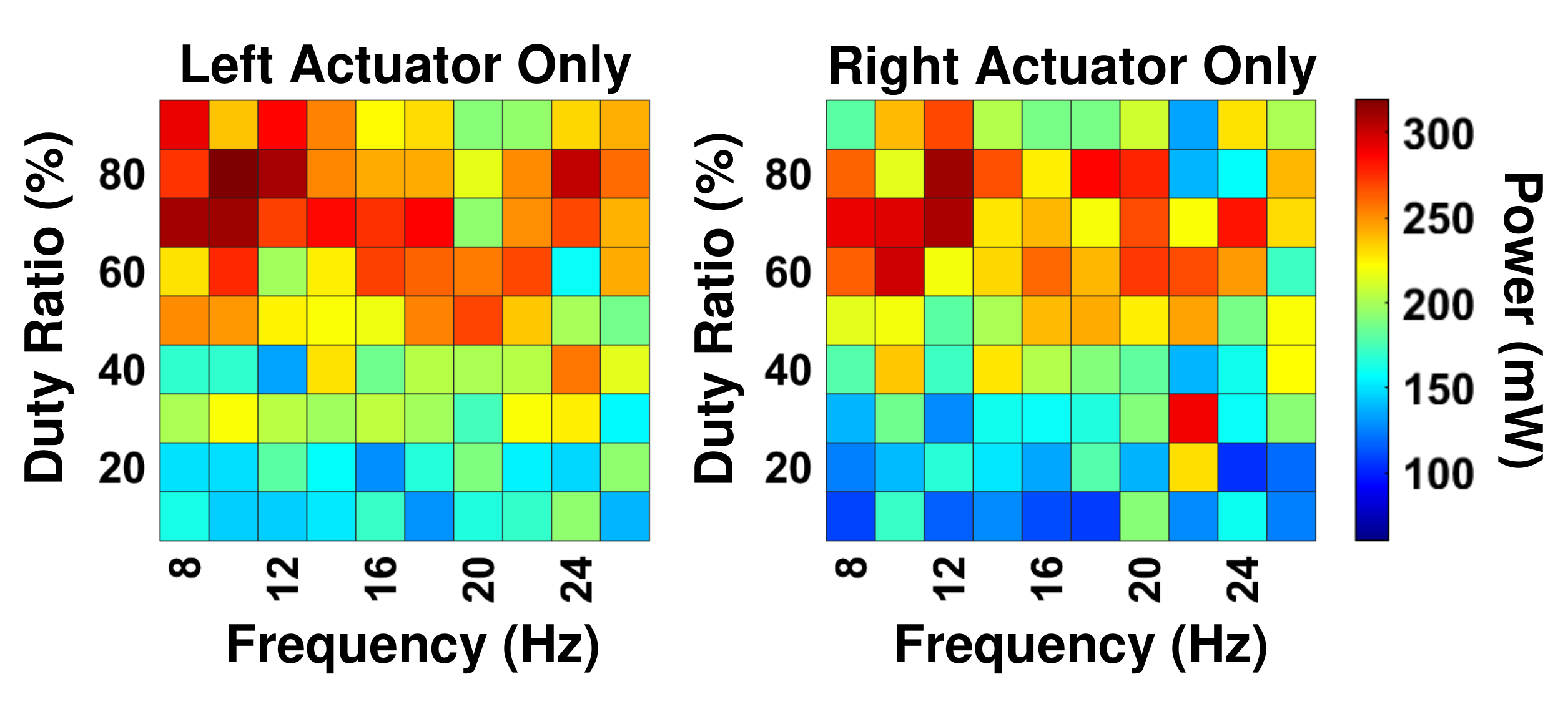}
\vspace{-18pt}
\caption{The power consumption of the soft robot power stage (power electronics and actuators) as a function of frequency and duty ratio when only driving the left actuator or the right actuator. Batteries at position \#1.}
\label{fig:BatPos1_PowerConsumption}
\vspace{-10pt}
\end{figure}

A test platform was constructed to collect data for experimental verification. The robot is tested on a 37$\times$6-inch marble tile (friction coefficient measured as 0.36). A fence around the experimental area enables IR distance sensing with 1~mm precision. The eViper calculates its velocity and current consumption and communicates with a computer through Bluetooth to transmit the data, enabling fully untethered operation. For each actuation pattern, the robot was activated for 5 seconds to ensure that it reaches a periodic steady state before the data was measured and transmitted back to the computer. The experiments start by actuating a single actuator with different frequencies (8~Hz to 26~Hz) and duty ratios (10\% to 90\%). The two actuators are then actuated with different frequencies (8~Hz to 26~Hz), phases (0$\degree$ to 324$\degree$), and duty ratios (0\% to 90\%). As shown in Fig.~\ref{fig:BatPosAll}, experiments are also conducted with the batteries placed at three different locations to study the weight impact on robot velocities. In the following sections, we define velocity as positive when the robot moves to the left, and negative when it moves to the right.

\subsection{Velocity and Efficiency when Actuating a Single Actuator}\label{LinearSpeedSingleAct}

\begin{figure}[t]
\centering
\includegraphics[width=\columnwidth]{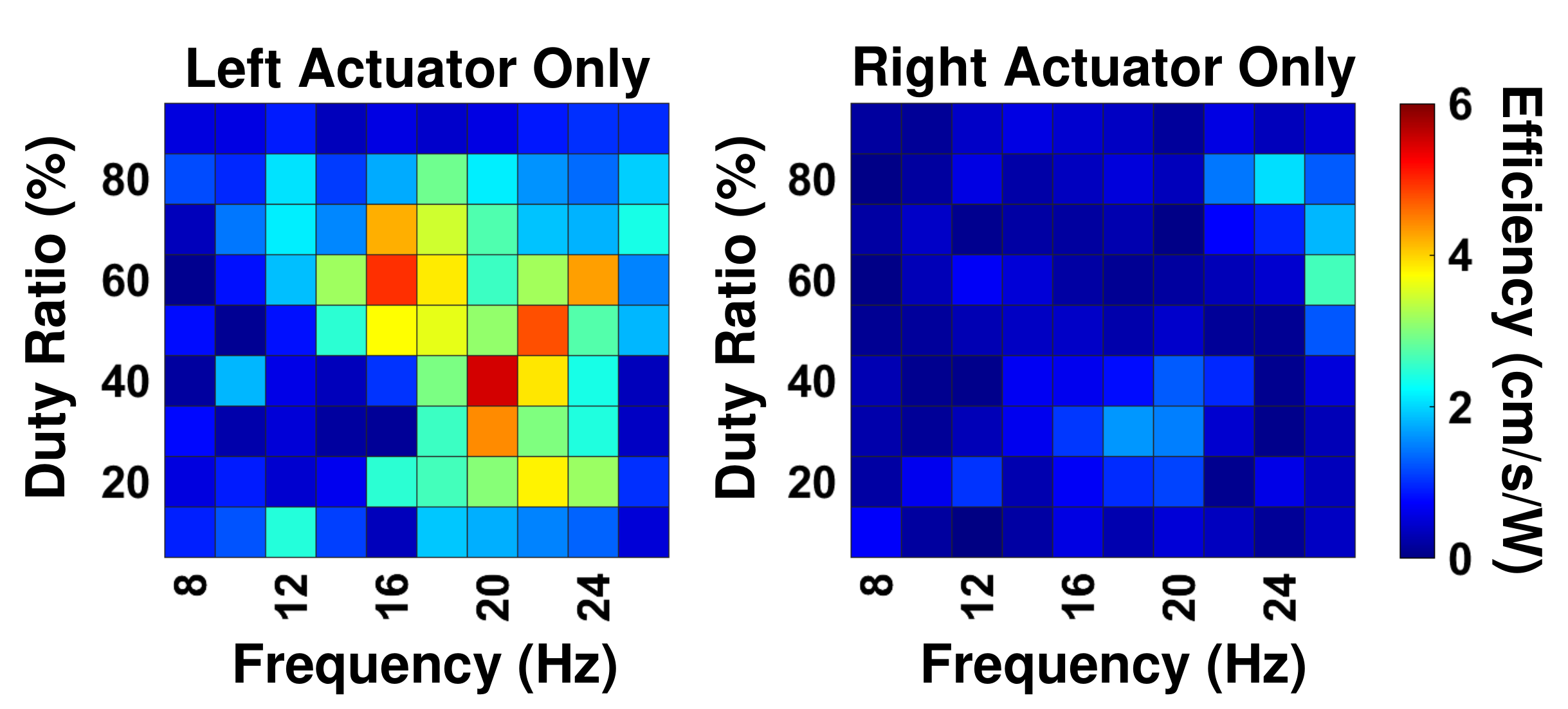}
\vspace{-18pt}
\caption{The actuation efficiency of the robot as a function of frequency and duty ratio when driving only the left actuator (left figure) or the right actuator (right figure). Driving the left actuator is more efficient than driving the right actuator. Carrying the batteries makes the right actuator less efficient. Batteries at position \#1.}
\label{fig:BatPos1_Eff}
\vspace{5pt}
\includegraphics[width=\columnwidth]{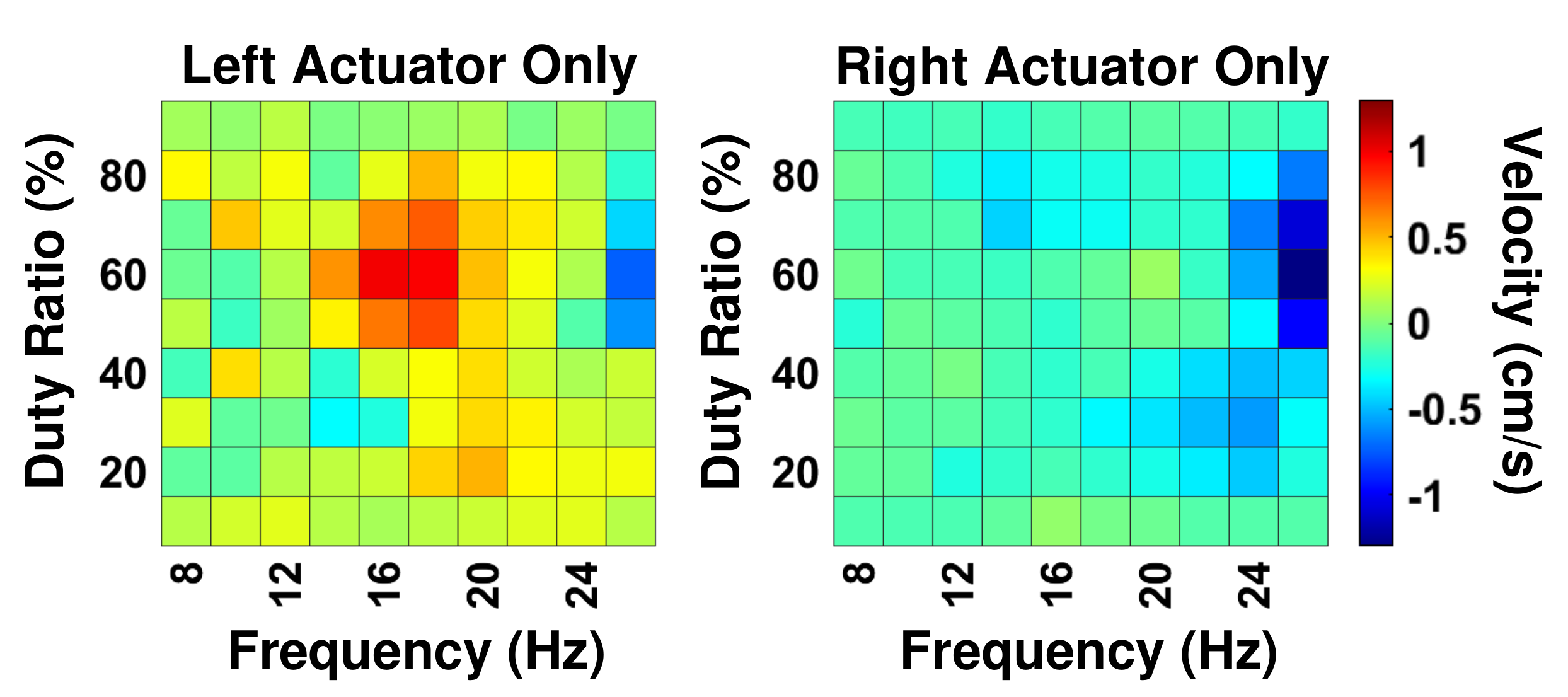}
\vspace{-18pt}
\caption{Experimentally measured soft robot velocity when driving only the left actuator (left figure) or the right actuator (right figure), with the batteries placed at position \#2. The left actuator is again more efficient.}
\label{fig:BatPos2}
\vspace{10pt}
\includegraphics[width=\columnwidth]{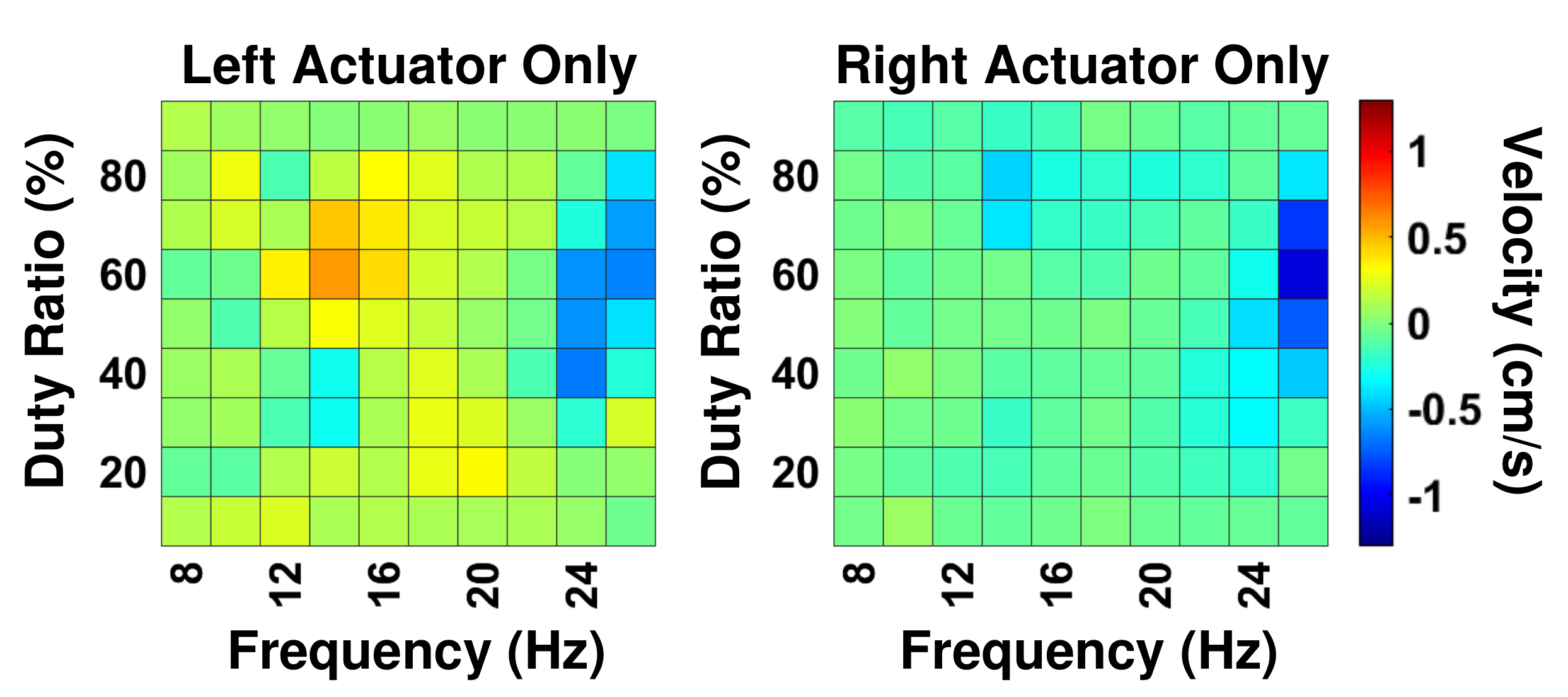}
\vspace{-18pt}
\caption{Experimentally measured velocity when driving only the left actuator (left figure) or the right actuator (right figure). The batteries are placed at position \#3. Both left and right actuators become less efficient.}
\label{fig:BatPos3}
\vspace{-10pt}
\end{figure}

This section presents the simulation and experimental results for driving a single actuator with different frequencies and duty ratios. The velocity and power consumption are measured. The batteries are placed at position \#1.

Figure~\ref{fig:BatPos1} shows and compares the simulated and measured velocity. In experiments, we measured a maximum positive velocity of 1.3~cm/s at 16~Hz and 60\% duty ratio when driving the left actuator only, and a maximum negative velocity of $-$0.5~cm/s at 26~Hz and 60\% duty ratio when driving the right actuator only. The simulation result for the left actuator shows good agreement with the experimental results. However, discrepancies between experiment and simulation are observed for the right actuator. This could be a consequence of the batteries sitting on the right actuator, where they may alter the mechanical properties in a way that is not captured by the simulation.

Figure~\ref{fig:BatPos1_PowerConsumption} depicts the power consumption for driving a single actuator at various frequency and duty ratio combinations. The values in the figure represent the power consumption of the power stage, which is the power consumed by the driver power electronics and the actuator. It can be estimated by multiplying the measured average current by 7.4~V (the battery voltage) and subtracting the computational power (about 740~mW). The power consumption is generally higher with larger duty ratios. This is related to our power electronics design -- the power consumption is linearly proportional to the duty ratio of the actuation pattern. A different implementation of the power electronics may lead to very different power consumption.

Dividing the absolute velocity (Fig.~\ref{fig:BatPos1}) by power (Fig.~\ref{fig:BatPos1_PowerConsumption}), we can estimate the locomotion efficiency in the unit of cm/s/W, as shown in Fig.~\ref{fig:BatPos1_Eff}. When driving the left actuator only, the efficiency is generally higher at mid-range frequencies and duty ratios. When driving the right actuator only, the efficiency is much lower due to the battery position.

\subsection{Impact of Battery Location on Soft Robot Locomotion}

In this section, we show the experimental results when moving batteries from position \#1 to position \#2 and position \#3. Only one actuator is actuated at a time.

Figures~\ref{fig:BatPos2}--\ref{fig:BatPos3} depicts the velocity measured in experiments. We can observe an obvious difference by comparing these two figures to Fig.~\ref{fig:BatPos1}. With the same actuation pattern, the velocities with batteries in positions \#2 and \#3 are greatly reduced from that in position \#1. These results indicate that the weight distribution of components has a significant impact on soft robot locomotion and must be considered when designing the soft robot body and placing the actuators.

The power consumption in positions \#2 and \#3 is similar to that in position \#1 (Fig.~\ref{fig:BatPos1_PowerConsumption}), because the electrical circuits work similarly regardless of the weight distribution, albeit with different mechanical responses and locomotion. Actuation efficiency can be computed in the same way. Due to a generally lower speed at positions \#2 and \#3, the efficiencies are also lower than those at position \#1 (Fig.~\ref{fig:BatPos1_Eff}).

\begin{figure}[t]
\centering
\includegraphics[width=\columnwidth]{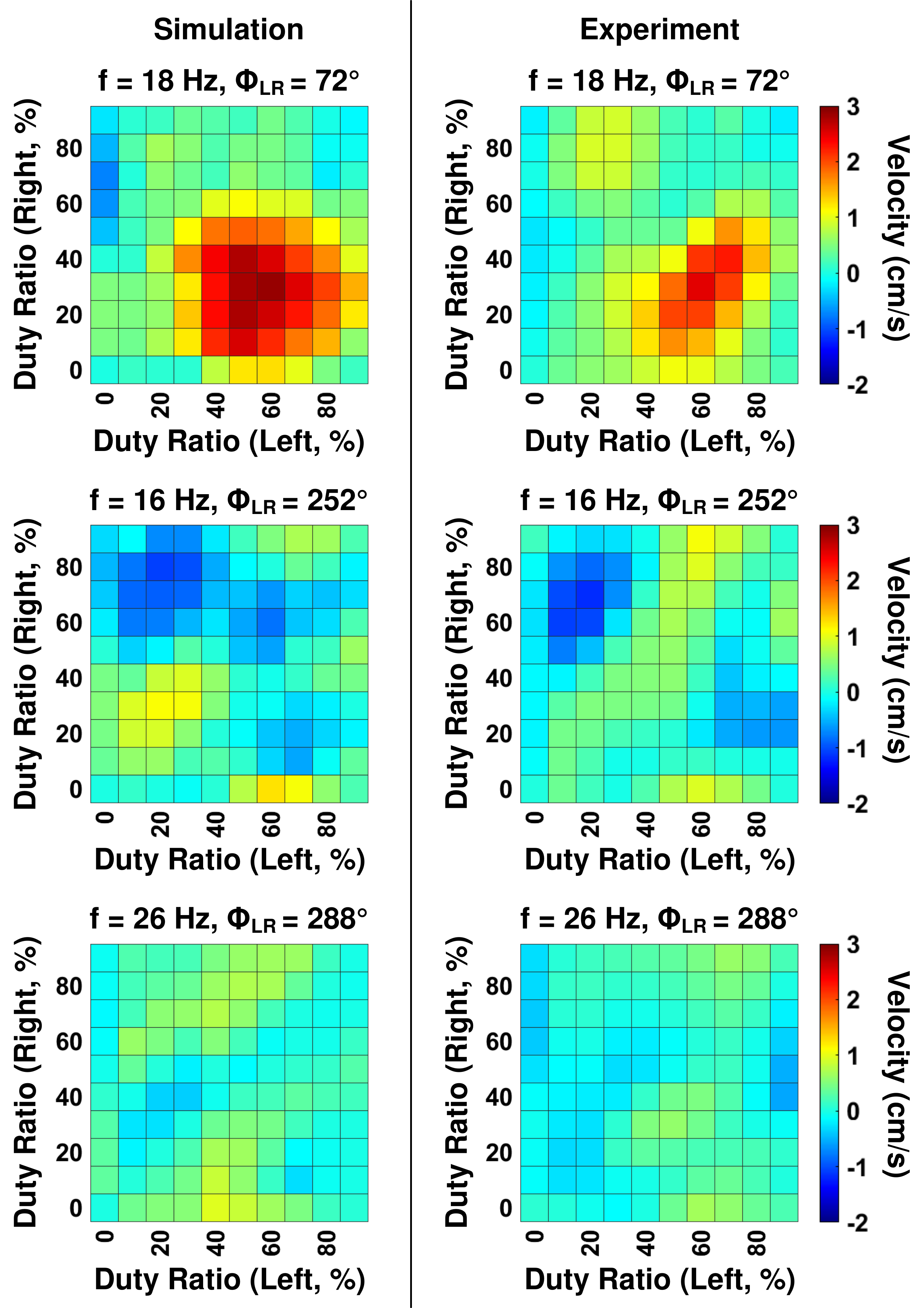}
\vspace{-18pt}
\caption{Simulated (left) and measured (right) soft robot velocity when both actuators are activated. The two actuators are driven at the same frequency $f$ and a specific phase $\Phi_{LR}$, as shown in the title of each graph, but at different duty ratios. Batteries at position \#1.}
\label{fig:TwoActuatorCompareLinearSpeed}
\vspace{-10pt}
\end{figure}

\subsection{Velocity and Efficiency when Driving Two Actuators}

\begin{figure}[t]
\centering
\includegraphics[width=\columnwidth]{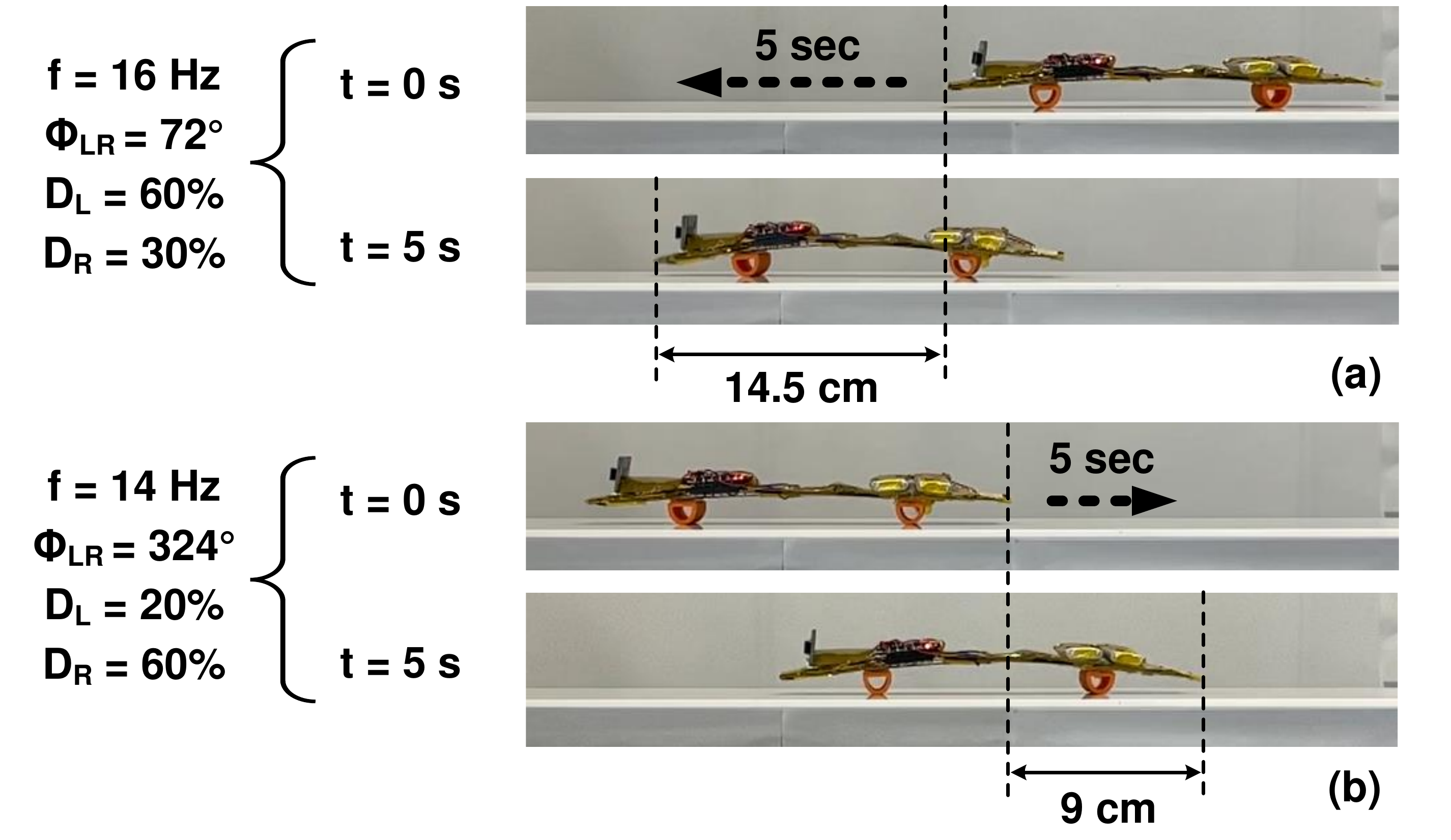}
\vspace{-15pt}
\caption{(a) eViper reaches 2.9 cm/s when operated at the maximum efficiency point of positive velocity. (b) eViper reaches -1.8 cm/s when operated at the maximum efficiency point of negative velocity.}
\label{fig:VideoScreenshots}
\vspace{5pt}
\includegraphics[width=\columnwidth]{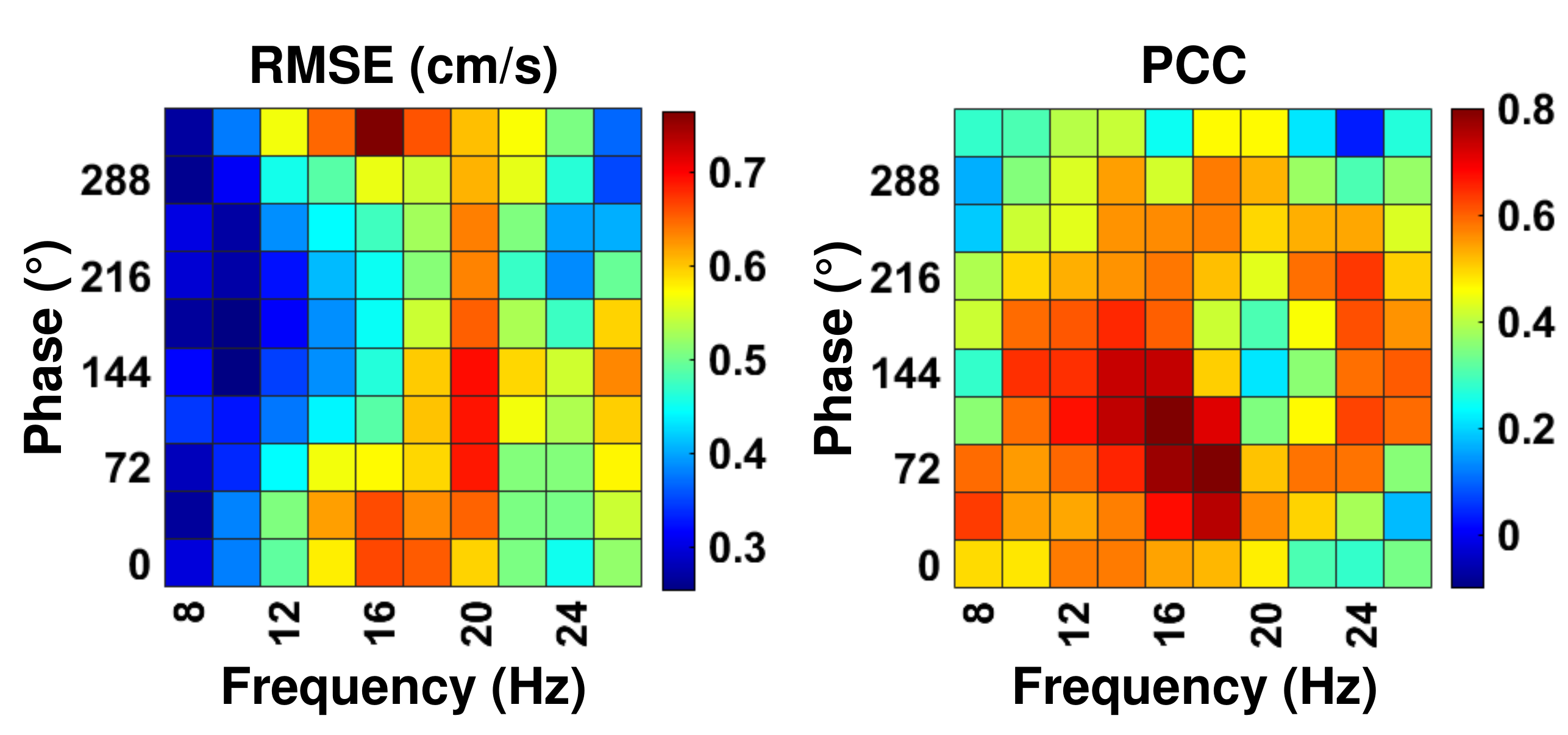}
\vspace{-15pt}
\caption{The root-mean-square error (RMSE) and Pearson Correlation Coefficient (PCC) of the simulated and measured velocity as a function of operating frequency and phase (for two-actuator operation).}
\label{fig:CorrelationAndRMSE}
\vspace{10pt}
\includegraphics[width=\columnwidth]{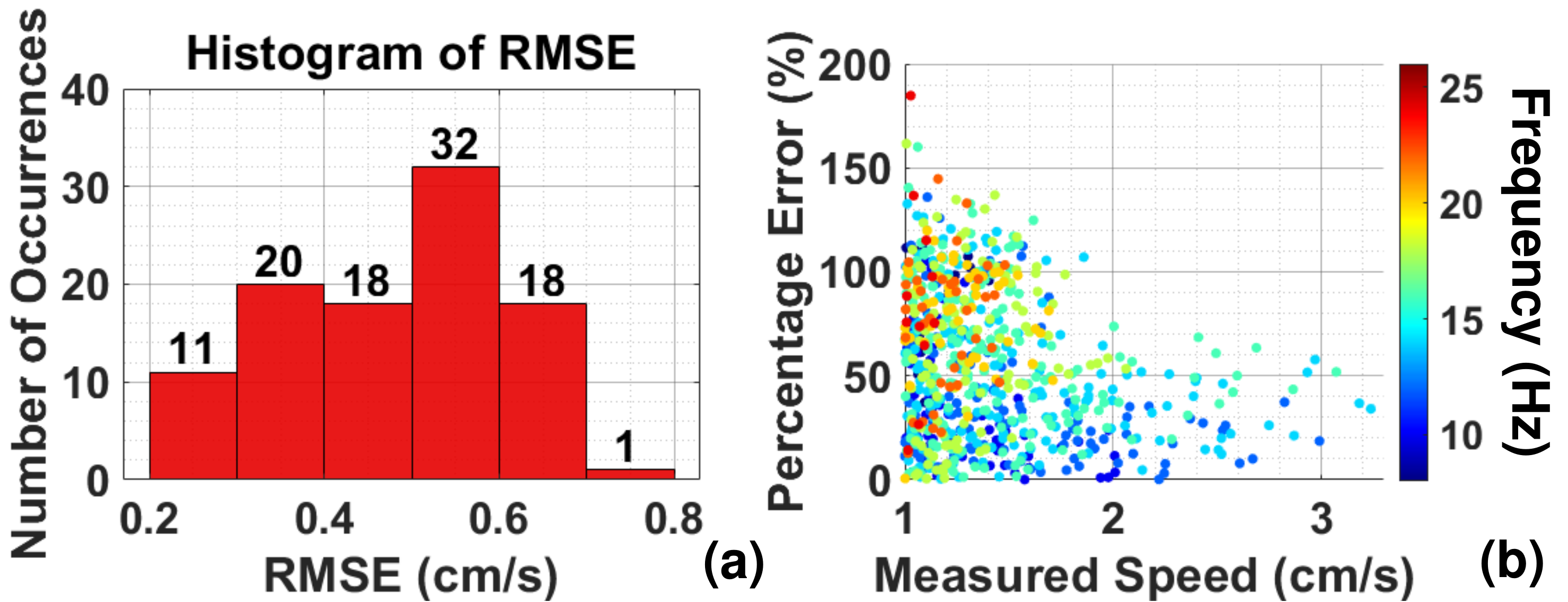}
\vspace{-15pt}
\caption{(a) Histogram of the root mean square error (RMSE) shown in Fig.~\ref{fig:CorrelationAndRMSE}. (b) Percentage error of the absolute velocity. The model is less accurate when the robot is moving slow (slower than 2 cm/s) and more accurate when the robot is moving fast (faster than 2 cm/s).}
\label{fig:Histograms}
\vspace{-10pt}
\end{figure}

In this section, we compare the data obtained in simulation and experiments when driving two actuators together. The batteries are at position \#1 for all data presented. To ensure periodic steady-state operation, both actuators are actuated at the same frequency. We sweep the operating frequency together with the phase between the two actuators and the duty ratio of each actuator, and compare the locomotion results of simulations and experiments.

Figure~\ref{fig:TwoActuatorCompareLinearSpeed} shows that in both simulation and experiments, the peak positive and negative velocity usually is reached at mid-range frequencies (14~Hz to 18~Hz), and that more noise is observed at higher and lower frequencies. The data are drawn with respect to the left- and right-actuator duty ratios at a fixed frequency and phase. In the following paragraphs, we call this a frequency-phase combination. As presented in \cite{ZhiwuICRA}, the piezoelectric soft robot is best operated close to the resonant frequency of the actuators. In general, driving the actuators at mid-range frequencies, low phase difference, high left-actuator and low right-actuator duty ratios leads to high velocity toward the left. Driving the actuators at mid-range frequencies, high phase difference, low left-actuator and high right-actuator duty ratios leads to high velocity toward the right. SFERS and experiments agree well on these characteristics. No clear trends are observed at lower or higher frequencies, for example in the third row of Fig.~\ref{fig:TwoActuatorCompareLinearSpeed}. 

The power consumption of a two-actuator eViper is generally proportional to the summation of their duty ratios. The actuation efficiency can be derived with the same approach. When the robot moves left, the maximum efficiency is 9.5~cm/s/W, obtained at 16~Hz, 72$\degree$ phase, 60\% left duty ratio, and 30\% right duty ratio. When the robot moves right, the maximum efficiency is 6.6~cm/s/W, which is achieved at 14~Hz, 324$\degree$ phase, 20\% left duty ratio, and 60\% right duty ratio. Figure~\ref{fig:VideoScreenshots} shows the motion of the robot operating at the two operation points.

\begin{figure}[t]
\centering
\includegraphics[width=\columnwidth]{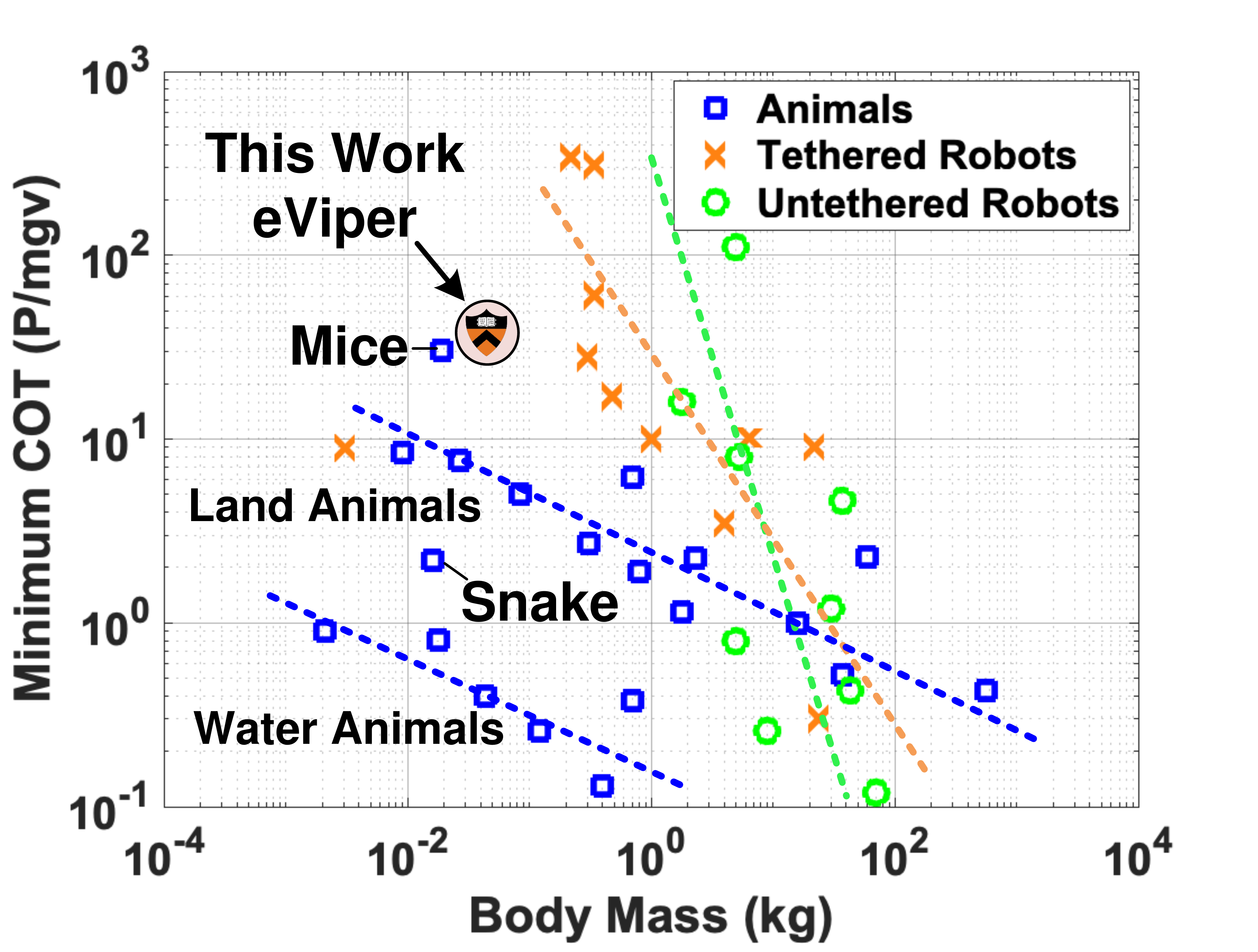}
\caption{Cost of transport (COT) of the eViper compared to other published soft or rigid robot systems. eViper is much lighter than most other untethered robot designs. The lowest COT of the soft eViper is about 36.4~W/(kg$\cdot$m$^2$/s$^3$), which is similar to the COT of mice (data from \cite{Baines2022}). The dotted lines depict the fitted mass-COT trends for land animals, water animals, tethered robots, and untethered robots.}
\label{fig:COT}
\vspace{-10pt}
\end{figure}

\subsection{Simulation Error Analysis}
Root-mean-square error (RMSE) and Pearson Correlation Coefficient (PCC) are used to compare the simulation and experimental results at each frequency-phase combination. They are visualized as 2-D heat maps in Fig.~\ref{fig:CorrelationAndRMSE}. RMSE describes the absolute error between the simulated and measured velocity. PCC represents the correlation as a means of quantifying the similarity in shape of the velocity contour between simulation and measurement. RMSE and PCC jointly quantifies the applicability and limitations of SFERS. The RMSE is defined as:
\begin{equation}
RMSE=\sqrt{\frac{1}{N}\sum_{i=1}^{N} (X_i-Y_i)^2}.
\end{equation}
\noindent The PCC is defined as:
\begin{equation}
PCC=\frac{\sum_{i=1}^{N}(X_i-\overline{X})(Y_i-\overline{Y})}{\sqrt{(\sum_{i=1}^{N}(X_i-\overline{X})^2)(\sum_{i=1}^{N}(Y_i-\overline{Y})^2)}}.
\end{equation}
where $i$ is the index of the data, $N$ is the total amount of data in each frequency-phase combination, $X_i$ and $Y_i$ are the value of the simulation and experimental data at a specific index, and $\overline{X}$ and $\overline{Y}$ are the average values of the simulation and experimental results in a frequency-phase combination. 

The result is presented in Fig.~\ref{fig:CorrelationAndRMSE}. For example, comparing the data in the top row of Fig.~\ref{fig:TwoActuatorCompareLinearSpeed} results in an RMSE of 0.59~cm/s and a correlation coefficient of 0.8. The RMSE graph tells us that the errors at mid-range frequencies are relatively high. This could result from the fact that the velocities at mid-range frequencies have a larger absolute value, or it could also come from the simulation errors discussed in Section~\ref{SFERS}. At the right half of Fig.~\ref{fig:CorrelationAndRMSE}, there is a ``ring-shaped" region in the middle of the graph where the PCCs are larger than 0.5, and the values at the corners and the center of the ring are less satisfactory. Figure~\ref{fig:Histograms}(a) shows the histogram of the RMSE. RMSEs are generally below 0.8 cm/s. Figure~\ref{fig:Histograms}(b) says that the model is less accurate when the robot is moving slow (slower than 2 cm/s) and more accurate when the robot is moving fast (faster than 2 cm/s).

\subsection{Comparisons with Animals and Other Robots}
A common figure-of-merit (FOM) for comparing the locomotion efficiency is the cost of transport (COT) \cite{Baines2022}, COT $= \frac{P}{mgv}$, where $P$ is the power consumption, $m$ is the robot mass, $g$ is the gravitational acceleration, and $v$ is the robot speed. For the two-actuator eViper, the most energy-efficient movement pattern is obtained when $P=0.46$~W, $m = 0.0445$~kg, $g = 9.8$~m/s$^2$, and $v = 0.029$~m/s. The COT of the eViper is about 36.4 W/(kg$\cdot$m$^2$/s$^3$). As shown in Fig.~\ref{fig:COT}, the weight of eViper is lower than most other robots. Its COT is comparable to land animals (mice).

\subsection{Future Work}
The current eViper prototype does not have closed-loop control onboard. Rotational locomotion is also needed for the robot to crawl in a 2D space. The project will extend the geometry of eViper from 1D to 2D (a cross-shaped robot) and add closed-loop control functionalities onboard to gain precise control on both lateral and rotational motion. Another direction is to install strain sensors on actuators and predict the velocity of eViper by processing the sensor signals.

\section{Conclusion}

This paper presents eViper and SFERS as an integrated platform for studying the impact of actuation patterns and weight distribution on soft robot locomotion. eViper is a scalable untethered soft robot with embedded power, energy storage, sensing, and computing. SFERS is a software toolbox based on PyBullet that can predict the general movement pattern of multi-actuator soft robots. The two-actuator eViper was built and thoroughly tested by activating individual or both actuators. It achieves a maximum speed of 3~cm/s and a minimum COT of 36.4~W/(kg$\cdot$m$^2$/s$^3$).

\section*{Acknowledgment}
This work was jointly supported by Semiconductor Research Corporation, Princeton Materials Institute, and Princeton Andlinger Center for Energy and the Environment.

\bibliographystyle{IEEEtran}
\bibliography{IEEEabrv, repeat_name, bibliography, extra_bib}

\begin{thebibliography}{10}
\providecommand{\url}[1]{#1}
\csname url@samestyle\endcsname
\providecommand{\newblock}{\relax}
\providecommand{\bibinfo}[2]{#2}
\providecommand{\BIBentrySTDinterwordspacing}{\spaceskip=0pt\relax}
\providecommand{\BIBentryALTinterwordstretchfactor}{4}
\providecommand{\BIBentryALTinterwordspacing}{\spaceskip=\fontdimen2\font plus
\BIBentryALTinterwordstretchfactor\fontdimen3\font minus
  \fontdimen4\font\relax}
\providecommand{\BIBforeignlanguage}[2]{{%
\expandafter\ifx\csname l@#1\endcsname\relax
\typeout{** WARNING: IEEEtran.bst: No hyphenation pattern has been}%
\typeout{** loaded for the language `#1'. Using the pattern for}%
\typeout{** the default language instead.}%
\else
\language=\csname l@#1\endcsname
\fi
#2}}
\providecommand{\BIBdecl}{\relax}
\BIBdecl

\bibitem{Rich2018}
\BIBentryALTinterwordspacing
S.~I. Rich, R.~J. Wood, and C.~Majidi, ``Untethered soft robotics,''
  \emph{Nature Electronics}, vol.~1, no.~2, pp. 102--112, Feb 2018. [Online].
  Available: \url{https://doi.org/10.1038/s41928-018-0024-1}
\BIBentrySTDinterwordspacing

\bibitem{Li2022}
\BIBentryALTinterwordspacing
M.~Li, A.~Pal, A.~Aghakhani, A.~Pena-Francesch, and M.~Sitti, ``Soft actuators
  for real-world applications,'' \emph{Nature Reviews Materials}, vol.~7,
  no.~3, pp. 235--249, Mar 2022. [Online]. Available:
  \url{https://doi.org/10.1038/s41578-021-00389-7}
\BIBentrySTDinterwordspacing

\bibitem{Rus2015}
\BIBentryALTinterwordspacing
D.~Rus and M.~T. Tolley, ``Design, fabrication and control of soft robots,''
  \emph{Nature}, vol. 521, no. 7553, pp. 467--475, May 2015. [Online].
  Available: \url{https://doi.org/10.1038/nature14543}
\BIBentrySTDinterwordspacing

\bibitem{Picardi17}
\BIBentryALTinterwordspacing
M.~Calisti, G.~Picardi, and C.~Laschi, ``Fundamentals of soft robot
  locomotion,'' \emph{Journal of The Royal Society Interface}, vol.~14, no.
  130, p. 20170101, 2017. [Online]. Available:
  \url{https://royalsocietypublishing.org/doi/abs/10.1098/rsif.2017.0101}
\BIBentrySTDinterwordspacing

\bibitem{iida2011soft}
F.~Iida and C.~Laschi, ``Soft robotics: Challenges and perspectives,''
  \emph{Procedia Computer Science}, vol.~7, pp. 99--102, 2011.

\bibitem{laschi2016soft}
C.~Laschi, B.~Mazzolai, and M.~Cianchetti, ``Soft robotics: Technologies and
  systems pushing the boundaries of robot abilities,'' \emph{Science robotics},
  vol.~1, no.~1, p. eaah3690, 2016.

\bibitem{wu2019insect}
Y.~Wu, J.~K. Yim, J.~Liang, Z.~Shao, M.~Qi, J.~Zhong, Z.~Luo, X.~Yan, M.~Zhang,
  X.~Wang \emph{et~al.}, ``Insect-scale fast moving and ultrarobust soft
  robot,'' \emph{Science robotics}, vol.~4, no.~32, p. eaax1594, 2019.

\bibitem{Jafferis2019}
\BIBentryALTinterwordspacing
N.~T. Jafferis, E.~F. Helbling, M.~Karpelson, and R.~J. Wood, ``Untethered
  flight of an insect-sized flapping-wing microscale aerial vehicle,''
  \emph{Nature}, vol. 570, no. 7762, pp. 491--495, Jun 2019. [Online].
  Available: \url{https://doi.org/10.1038/s41586-019-1322-0}
\BIBentrySTDinterwordspacing

\bibitem{elastomer19}
\BIBentryALTinterwordspacing
X.~Ji, X.~Liu, V.~Cacucciolo, M.~Imboden, Y.~Civet, A.~E. Haitami, S.~Cantin,
  Y.~Perriard, and H.~Shea, ``An autonomous untethered fast soft robotic insect
  driven by low-voltage dielectric elastomer actuators,'' \emph{Science
  Robotics}, vol.~4, no.~37, p. eaaz6451, 2019. [Online]. Available:
  \url{https://www.science.org/doi/abs/10.1126/scirobotics.aaz6451}
\BIBentrySTDinterwordspacing

\bibitem{footpad21}
\BIBentryALTinterwordspacing
J.~Liang, Y.~Wu, J.~K. Yim, H.~Chen, Z.~Miao, H.~Liu, Y.~Liu, Y.~Liu, D.~Wang,
  W.~Qiu, Z.~Shao, M.~Zhang, X.~Wang, J.~Zhong, and L.~Lin, ``Electrostatic
  footpads enable agile insect-scale soft robots with trajectory control,''
  \emph{Science Robotics}, vol.~6, no.~55, p. eabe7906, 2021. [Online].
  Available: \url{https://www.science.org/doi/abs/10.1126/scirobotics.abe7906}
\BIBentrySTDinterwordspacing

\bibitem{el2020soft}
N.~El-Atab, R.~B. Mishra, F.~Al-Modaf, L.~Joharji, A.~A. Alsharif, H.~Alamoudi,
  M.~Diaz, N.~Qaiser, and M.~M. Hussain, ``Soft actuators for soft robotic
  applications: a review,'' \emph{Advanced Intelligent Systems}, vol.~2,
  no.~10, p. 2000128, 2020.

\bibitem{Efficiency17}
\BIBentryALTinterwordspacing
L.~Shui, L.~Zhu, Z.~Yang, Y.~Liu, and X.~Chen, ``Energy efficiency of mobile
  soft robots,'' \emph{Soft Matter}, vol.~13, pp. 8223--8233, 2017. [Online].
  Available: \url{http://dx.doi.org/10.1039/C7SM01617D}
\BIBentrySTDinterwordspacing

\bibitem{Zheng2021}
\BIBentryALTinterwordspacing
Z.~Zheng, P.~Kumar, Y.~Chen, H.~Cheng, S.~Wagner, M.~Chen, N.~Verma, and J.~C.
  Sturm, ``{Piezoelectric Soft Robot Inchworm Motion by Controlling Ground
  Friction through Robot Shape},'' nov 2021. [Online]. Available:
  \url{http://arxiv.org/abs/2111.00944}
\BIBentrySTDinterwordspacing

\bibitem{ZhiwuRobo}
Z.~Zheng, P.~Kumar, Y.~Chen, H.~Cheng, S.~Wagner, M.~Chen, N.~Verma, and J.~C.
  Sturm, ``Model-based control of planar piezoelectric inchworm soft robot for
  crawling in constrained environments,'' in \emph{2022 IEEE 5th Inter. Conf.
  on Soft Robo. (RoboSoft)}, 2022.

\bibitem{ZhiwuICRA}
Z.~Zheng, P.~Kumar, Y.~Chen, H.~Cheng, S.~Wagner, M.~Chen, N.~Verma, and J.~C.
  Sturm, ``Scalable simulation and demonstration of jumping piezoelectric 2-d
  soft robots,'' in \emph{2022 International Conference on Robotics and
  Automation (ICRA)}, 2022, pp. 5199--5204.

\bibitem{yu2020crawling}
M.~Yu, W.~Yang, Y.~Yu, X.~Cheng, and Z.~Jiao, ``A crawling soft robot driven by
  pneumatic foldable actuators based on miura-ori,'' in \emph{Actuators},
  vol.~9, no.~2.\hskip 1em plus 0.5em minus 0.4em\relax MDPI, 2020, p.~26.

\bibitem{wu2018structure}
P.~Wu, W.~Jiangbei, and F.~Yanqiong, ``The structure, design, and closed-loop
  motion control of a differential drive soft robot,'' \emph{Soft robotics},
  vol.~5, no.~1, pp. 71--80, 2018.

\bibitem{ICRA16}
A.~S. Chen and S.~Bergbreiter, ``Electroadhesive feet for turning control in
  legged robots,'' in \emph{2016 IEEE International Conference on Robotics and
  Automation (ICRA)}, 2016, pp. 3806--3812.

\bibitem{calabrese2019soft}
L.~Calabrese, A.~Berardo, D.~De~Rossi, M.~Gei, N.~M. Pugno, and G.~Fantoni, ``A
  soft robot structure with limbless resonant, stick and slip locomotion,''
  \emph{Smart Materials and Structures}, vol.~28, no.~10, p. 104005, 2019.

\bibitem{MEMS19}
J.~Liang, Y.~Wu, Z.~Shao, J.~K. Yim, R.~Xu, Y.~Song, M.~Qi, J.~Zhong, M.~Zhang,
  X.~Wang, and L.~Lin, ``Manipulating the moving trajectory of insect-scale
  piezoelectric soft robots by frequency,'' in \emph{2019 IEEE 32nd
  International Conference on Micro Electro Mechanical Systems (MEMS)}, 2019,
  pp. 1041--1044.

\bibitem{Hassan16}
H.~H. Hariri, L.~A. Prasetya, S.~Foong, G.~S. Soh, K.~N. Otto, and K.~L. Wood,
  ``A tether-less legged piezoelectric miniature robot using bounding gait
  locomotion for bidirectional motion,'' in \emph{2016 IEEE Inter. Conf. on
  Robo. and Automation (ICRA)}, 2016, pp. 4743--4749.

\bibitem{coumans2016pybullet}
E.~Coumans and Y.~Bai, ``Pybullet, a python module for physics simulation for
  games, robotics and machine learning,'' 2016.

\bibitem{Smartmaterial}
\BIBentryALTinterwordspacing
``{Smart Material Corp. Sarasota, Florida. Part numbers: M-8514-P1,
  M-8514-P2}.'' [Online]. Available:
  \url{https://www.smart-material.com/media/Datasheets/MFC \textunderscore
  V2.4-datasheet-web.pdf}
\BIBentrySTDinterwordspacing

\bibitem{https://doi.org/10.48550/arxiv.2207.00658}
\BIBentryALTinterwordspacing
Z.~Zheng, H.~Cheng, P.~Kumar, S.~Wagner, M.~Chen, N.~Verma, and J.~C. Sturm,
  ``Wirelessly-controlled untethered piezoelectric planar soft robot capable of
  bidirectional crawling and rotation,'' 2022. [Online]. Available:
  \url{https://arxiv.org/abs/2207.00658}
\BIBentrySTDinterwordspacing

\bibitem{Hsin22}
H.~Cheng, Z.~Zheng, P.~Kumar, Y.~Chen, and M.~Chen, ``Hybrid-soro: Hybrid
  switched capacitor power management architecture for multi-channel
  piezoelectric soft robot,'' in \emph{2022 IEEE Applied Power Electronics
  Conference and Exposition (APEC)}, 2022, pp. 1338--1344.

\bibitem{Baines2022}
\BIBentryALTinterwordspacing
R.~Baines, S.~K. Patiballa, J.~Booth, L.~Ramirez, T.~Sipple, A.~Garcia,
  F.~Fish, and R.~Kramer-Bottiglio, ``Multi-environment robotic transitions
  through adaptive morphogenesis,'' \emph{Nature}, vol. 610, no. 7931, pp.
  283--289, Oct 2022. [Online]. Available:
  \url{https://doi.org/10.1038/s41586-022-05188-w}
\BIBentrySTDinterwordspacing

\end{thebibliography}
\pagebreak
\end{document}